\documentclass[journal]{IEEEtran}
\usepackage{multicol, blindtext}
\usepackage[english]{babel}
\usepackage{comment}
\usepackage{graphicx}  %Required
\usepackage{amssymb,amsfonts}
\usepackage[fleqn]{amsmath}
\usepackage{subfig}
\usepackage{url}

\newcommand{\method}{$MRFusion$}

\begin{document}
\title{MRFusion: A Deep Learning architecture to fuse PAN and MS imagery for land cover mapping}

\author{R. Gaetano,
        D. Ienco,
        K. Ose,
        R. Cresson

\thanks{R. Gaetano is with CIRAD, UMR TETIS, 500 Rue J.-F. Breton, F-34000 Montpellier, France and with UMR TETIS, Univ. Montpellier, AgroParisTech, CIRAD, CNRS, IRSTEA, Montpellier, France (email: raffaele.gaetano@cirad.fr).}
\thanks{R. Cresson, K. Ose and D. Ienco are with UMR-TETIS laboratory, IRSTEA, University of Montpellier, Montpellier, France (email: remi.cresson@irstea.fr; dino.ienco@irstea.fr; paola.benedetti@irstea.fr; kenji.ose@irstea.fr).}
\thanks{D. Ienco is with LIRMM laboratory, Montpellier, France.} }

\maketitle

\begin{abstract}
Nowadays, Earth Observation systems provide a multitude of heterogeneous remote sensing data. How to manage such richness leveraging its complementarity is a crucial challenge in modern remote sensing analysis. Data Fusion techniques deal with this point proposing method to combine and exploit complementarity among the different data sensors. 

Considering optical Very High Spatial Resolution (VHSR) images, satellites obtain both Multi Spectral (MS) and panchromatic (PAN) images at different spatial resolution. VHSR images are extensively exploited to produce land cover maps to deal with agricultural, ecological, and socioeconomic issues as well as assessing ecosystem status, monitoring biodiversity and providing inputs to conceive food risk monitoring systems. Common techniques to produce land cover maps from such VHSR images typically opt for a prior pansharpening of the multi-resolution source for a full resolution processing.

Here, we propose a new deep learning architecture to jointly use PAN and MS imagery for a direct classification without any prior image fusion or resampling process. By managing the spectral information at its native spatial resolution, our method, named \method{}, aims at avoiding the possible information loss induced by pansharpening or any other hand-crafted preprocessing. Moreover, the proposed architecture is suitably designed to learn non-linear transformations of the sources with the explicit aim of taking as much as possible advantage of the complementarity of PAN and MS imagery.% to improve land cover mapping classification performances.

Experiments are carried out on two-real world scenarios depicting large areas with different land cover characteristics. The characteristics of the proposed scenarios underline the applicability and the generality of our method in operational settings.
\end{abstract}


\section{Introduction}
\label{sec:intro}
With the recent technological advances in the field of Earth Observation (EO), a multitude of satellite data sources are nowadays available. Due to this data abundance, one of the main challenges in modern remote sensing is the joint exploitation of different sources of information (data fusion) with the aim to leverage as much as possible the complementarity among different sources~\cite{Schmitt17}.

Among the different data fusion challenges, the single-sensor data fusion, involving optical Very High Spatial Resolution (VHSR) imagery acquired at multiple resolutions, still takes an important place. For most VHSR optical imagery, sensors provide both Multi Spectral (MS) and panchromatic (PAN) images, with the spatial resolution of PAN images higher than that of MS images~\cite{LiuJZZZLYT18}, such as IKONOS (4m MS and 1m Pan images), DEIMOS-2 (4m MS and 1m PAN images), Pl\'{e}iades (2m MS and 0.5m PAN images) and SPOT6/7 (6m MS and 1.5m PAN images). Such kind of EO data has been devoted great attention in the last decade since it is extremely useful to produce fine-scale land cover and/or land use mapping~\cite{GeorganosGVLSW18}.

Land cover/Land Use maps are an important tool to monitor human and physical environment. Land cover maps provide crucial inputs to interpret the nature of geographical, agricultural, ecological, and socioeconomic systems, such as characterizing hydrogeologic response to land-cover change, assessing ecosystem status, understanding spatial patterns of biodiversity, mapping urban areas for health issues, developing land management policies and providing inputs to conceive food risk monitoring systems~\cite{BegueABBAFLLSV18}.

Common techniques to deal with multi-resolution information coming from the same sensor are related to the use of pansharpening~\cite{4505280,Colditz06}. The pansharpening process aims to "sharpen" a Multi Spectral image using a panchromatic (single band) image. Generally, the common classification pipeline of multi-resolution VHSR images involves two main steps: i) fuse the different information at the same resolution by means of a downsampling/upsampling or pansharpening procedure~\cite{Colditz06} and ii) classify the fused images by means of machine learning techniques~\cite{RegniersBLG16} or, more recently, Convolutional Neural Networks (CNNs)~\cite{MaggioriTCA17,VolpiT17}.
Considering this standard pipeline, performances can be affected by artifacts or noise introduced by the fusion process~\cite{Colditz06} since generic data fusion strategies are not conceived to support the classification or discrimination among land cover/land use classes. Unfortunately, only few techniques were proposed to directly manage multi-resolution classification avoiding the image fusion step and the artifacts introduced by this common strategy~\cite{WemmertPFG09,StorvikFS05,LiuJZZZLYT18}. 

The deep learning revolution~\cite{Zhang16} has shown that neural network models are well adapted tools for automatically managing and classifying remote sensing data. Due to the high performances obtained by these approaches on satellite image classification and retrieval~\cite{DLiuWZZHF18,GuoYZH18,LiZHZM18}, Deep Learning is becoming a common tool to manage and classify remote sensing data.

The main characteristic of this type of model is the ability to simultaneously extract features optimized for image classification and the associated classifier~\cite{BengioCV13}. In the Remote Sensing field, several works already have employed Deep Learning to classify remote sensing images or produce land cover mapping~\cite{DLiuWZZHF18,GuoYZH18,TianLXM18,ScottESMD17}. All such strategies adopt the common classification pipeline of multi-resolution VHSR images in which the learning model is fed by the fused image.
Differently from all previous deep learning approaches, \cite{LiuJZZZLYT18} recently proposes a new architecture, named $DMIL$, to cope with VHSR image land cover mapping dealing directly with Multi Spectral and Panchromatic information at their native resolutions. The $DMIL$ method upscales the Multi Spectral image to the same resolution of the Panchromatic one by means of deconvolution operators~\cite{VolpiT17,AudebertSL16,NohHH15} with the objective to work on PAN and MS information at the same (finest) resolution. Successively, it treats MS information as a flat vector ignoring the spatial information carried out by this source of data. 

In consonance with recent Remote sensing developments in the field of data fusion~\cite{LiuJZZZLYT18,XuLRDGZ18}, this paper presents a novel framework to cope with land cover mapping from VHSR images by Deep Learning. In order to exploit MS and PAN information at their native resolution, we propose a two-branch architecture named \method{}. Such model automatically combines information at different resolutions with the objective to maximize the discrimination among the different land cover classes and, simultaneously, avoiding any error-prone image fusion or upsampling.
Differently from~\cite{LiuJZZZLYT18}, we do not introduce any extra operation to upsample MS patches to the same resolution of PAN, and decide to exploit both the spatial and spectral information carried by MS bands as well as the fine spatial information supplied by the PAN source. 
The main contributions of this paper are summarized as follows:
\begin{itemize}
\item a new Neural Network architecture is proposed, which combines MS and PAN information at their native resolution through a end-to-end learning-from-scratch strategy;
\item the proposed approach is also exploited as feature extraction strategy to feed standard Machine Learning approaches with similar or even better classification performances;
\item an in-depth quantitative and qualitative analysis is presented with the aim of highlighting the benefit of the proposed approach for VHSR land cover mapping;
\item the proposed strategy is validated using two real-world land cover mapping scenarios over large geographical areas which exhibit contrasted landscape and environmental characteristics.
\end{itemize}

The rest of the article is organized as follows: Section~\ref{sec:method} introduces the Deep Learning Architecture for the data fusion between Panchromatic and Multi Spectral information. The study site and the associated data are presented in Section~\ref{sec:data}. The experimental setting and the evaluation are carried out and discussed in Section~\ref{sec:expe}. Finally, Section~\ref{sec:conclu} concludes.

\section{Method}
\label{sec:method}
\begin{figure}[ht!]
\centering
\includegraphics[width=.99\columnwidth]{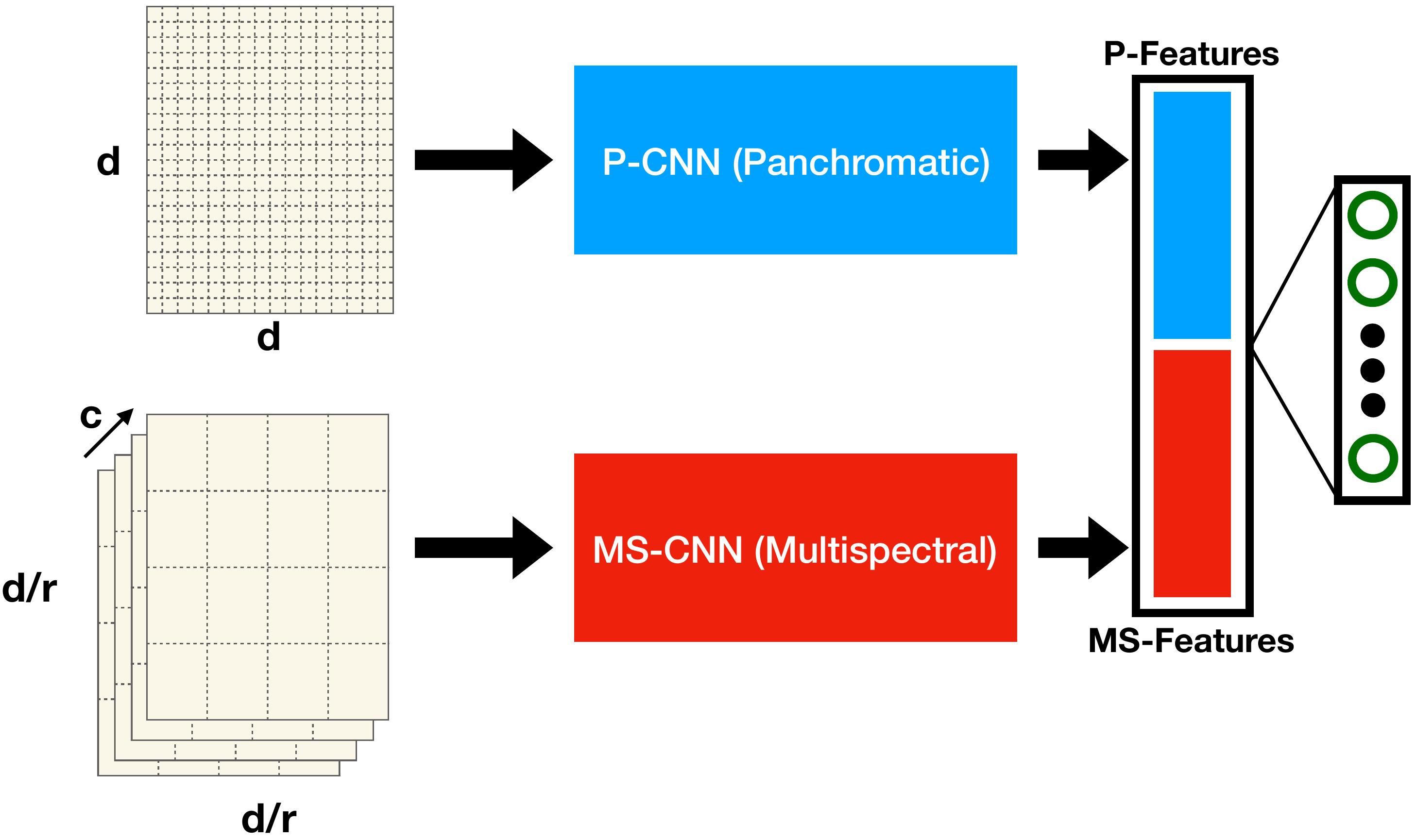}
\caption{ \label{fig:Overview} General Overview of \method. }
\end{figure}

In this Section we describe the proposed classification framework. Figure~\ref{fig:Overview} supplies a global overview of our proposal. 

Our Deep Learning architecture is composed of two parallel branches, independently processing PAN and MS patches of suitable sizes through a dedicated Convolutional Neural Network (CNN) module. It transforms the input source and produces a feature set that summarizes the corresponding spatial and spectral joint information. We name $P$-$CNN$ (resp. $MS$-$CNN$) the CNN working on the PAN image (resp. the MS images).
Downstream, the two features set are combined together by means of a concatenation and the whole set of features is successively used to perform the final classification. The Deep Learning Model is trained end-to-end from scratch.
Due to the different resolutions of the Panchromatic and Multi Spectral images, the two branches manage information at different spatial resolution. To ensure the generality of the network architecture choice, we have decided to input patches describing the same geographical zone into both branches of our architecture.
To this purpose, supposing that the spatial resolution ratio between the PAN and MS image is equal to $r$, we set the size of the PAN patches equals to ($d$ x $d$), hence a patch size of $d/r$ x $d/r$ for the MS image. More in detail, the $P$-$CNN$ branch takes as input a tensor of size $d$ x $d$ x 1 (since in the general case we only dispose of a single panchromatic band) where the parameter $d$ is used to define the patch size. Conversely, the branch associated to the $MS$-$CNN$ takes as input a tensor of size ($d/r$ x $d/r$ x $c$) where $c$ is the number of channels contained in the MS image.
We remind that, contrarily to~\cite{LiuJZZZLYT18} in which the MS image was upsampled inducing bias related to interpolation as well as increasing the amount of data to manage, \method{} directly deals with the different spatial resolutions avoiding additional interpolation biases and limiting the quantity of data to process. The prediction of \method{} is performed at the resolution of the PAN image. This means that our approach can be employed to produce land cover (or land use) maps at the finest spatial resolution among those of the input sources.
In the rest of this section we describe the Convolution Neural Networks ($MS$-$CNN$ and $P$-$CNN$) that are the core components of our framework. We also describe the training strategy we adopt to learn the parameters of our architecture.

\subsection{CNN architectures for the Panchromatic and the Multi Spectral information}

For both branches of our approach, we took inspiration from the VGG model~\cite{SimonyanZ14a}, one of the most well-known network architectures usually adopted to tackle with standard Computer Vision tasks. More in detail, for both branches we constantly increase the number of filters along the network as long as we reach a reasonable size of the feature maps.

\begin{figure}[ht!]
\centering
\includegraphics[width=.99\columnwidth]{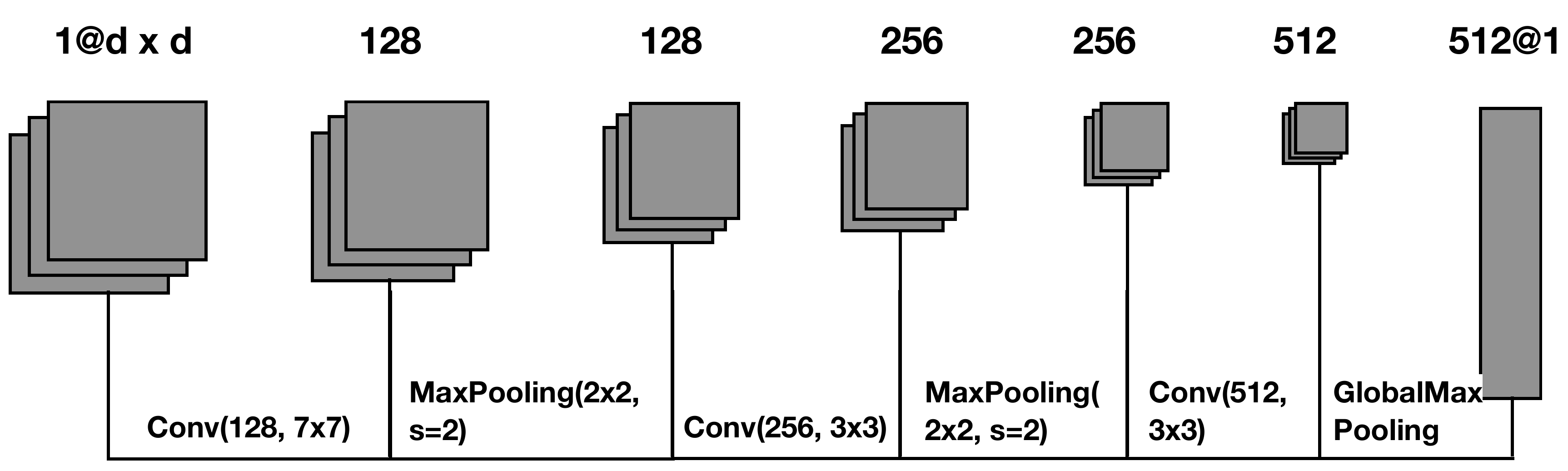}
\caption{ \label{fig:CNNPAN} $P$-$CNN$: Dedicated CNN Structure to manage Panchromatic information.}
\end{figure}

Considering the $P$-$CNN$ module (see Figure~\ref{fig:CNNPAN}), after each convolution we perform a Max Pooling operation with the aim of reducing the number of features to process and, at the same time, forcing the network module to focus on the most important part of the input signal. The first convolution has a kernel of 7$\times$7 and it produces 128 feature maps. The second and the third convolution have a kernel of 3$\times$3 and they produce respectively, 256 and 512 feature maps. At the end of the process, a Global Max Pooling is applied in order to extract 512 features (one feature for each feature maps obtained after the last convolution).

\begin{figure}[ht!]
\centering
\includegraphics[width=.99\columnwidth]{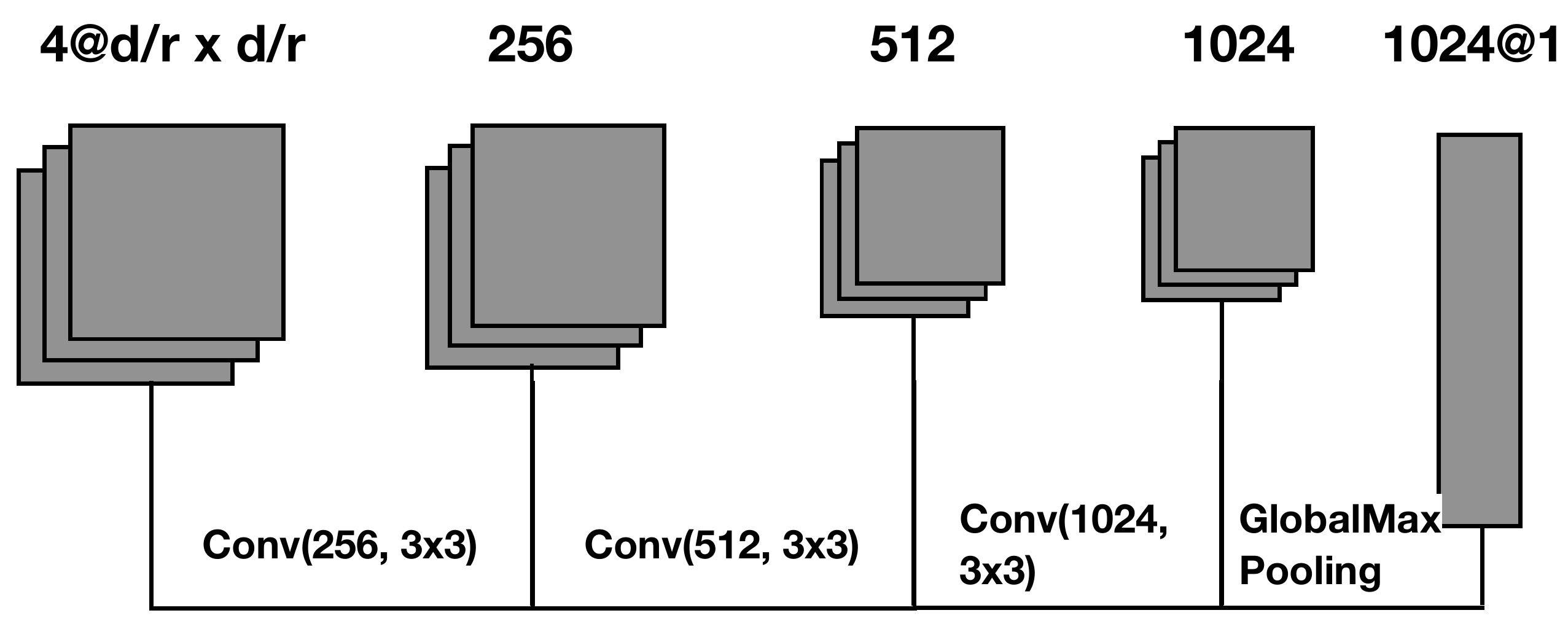}
\caption{ \label{fig:CNNMS} MS-CNN: Dedicated CNN Structure to manage  Panchromatic information.  }
\end{figure}

For the $MS$-$CNN$ module (Figure~\ref{fig:CNNMS}), no Max Pooling operation between two successive convolution stages is performed, for a better preservation of the spectral information. For the same reason, in each convolutional layer (three total layers as for $P$-$CNN$) the size of the kernel is limited to 3$\times$3. Moreover, each of these layers produces respectively 256, 512 and 1024 feature maps (doubled w.r.t. the corresponding $P$-$CNN$ layers) to deal with the richness of the spectral information and better exploit the correlations among the original MS bands. The final set of features summarizing the MS information is derived similarly to the $P$-$CNN$ model. Also in this case we apply a Global Max Pooling extracting 1024 features (one feature for each feature maps obtained after the last convolution).

%% RAF: j'enleve ca et j'explique les choix dans le texte precedent.
%Summarizing, the main differences between the two branches are the following:
%\begin{itemize}
%\item[1] The $P$-$CNN$ architecture has a first kernel of size 7$\times$7 while all the convolutional kernels of the $MS$-$CNN$ model have a size of 3$\times$3. Indeed, since the Multi Spectral patches are $r^2$ times smaller than the patches extracted from the PAN image, a 7$\times$7 kernel will compress too much the input information with the result of loosing informative signal portion. %\disc{I would rather explain that the physical size of MS patches are 4x larger than the PAN patches, hence small kernel are used to deal with the MS patches?}
%\item[2] We avoid Max Pooling operations in the $MS$-$CNN$ branch following the same principles we adopted before. %\disc{remove "following the same reasoning we had before"?} 
%Max Pooling layers can compress too much the information propagated to further level (too few convolutional layers) negatively affecting the discriminative power of the model. 
%\item[3] We decide to generate a bigger number of feature maps for the Multi Spectral branch than for the Panchromatic one according to the amount of spectral information supplied as input. Since the MS image contains more spectral richness w.r.t. the PAN image, a bigger number of feature maps will be more adequate to derive spectral correlation among the original channels.
%\end{itemize}

The features extracted by each branch of the architecture are successively merged by concatenation. Such set of concatenated features, 512 from the PAN branch and 1024 from the MS branch for a total of 1536 feature descriptors, is directly fully connected to the output layer in order to perform the final classification. The SoftMax activation function is finally applied~\cite{Zhang16} on the output layer with the aim to produce a kind of probability distribution over the class labels. The model weights are learned by back-propagation.

In both branches, each convolution is associated with a linear filter, followed by a Rectifier Linear Unit (ReLU) activation function~\cite{NairH10} to induce non-linearity and a batch normalization step~\cite{IoffeS15}. The ReLU activation function is defined as follows:
\begin{align}
ReLU(x) &= Max(0, W \cdot x + b) \label{eqn:relu}
\end{align}

This activation function is defined on the positive part of the linear transformation of its argument ($W \cdot x + b$).  The choice of ReLU nonlinearities is motivated by two factors: i) the good convergence properties it guarantees and ii) the low computational complexity it provides~\cite{NairH10}. Furthermore, batch normalization~\cite{IoffeS15} accelerates Deep Network training convergence by reducing the internal covariate shift.

\subsection{Network Training Strategy }
Due to the network architecture peculiarity (two branches, multi-scale input, different number of channels for branch) we learn the network weights end-to-end from scratch since we cannot reuse any existing available pre-trained architecture.
The cost function associated to our model is :
\begin{align}
LOSS &= L([PAN_{feat}, MS_{feat}], W, b) \label{eqn:cost}
\end{align}

where 
\begin{align}
L(feat, W, b) &= L( Y, SoftMax(feat \cdot W + b)) \nonumber
\end{align}
with $Y$ being the true value of the class variable. 
The cost function is modeled through categorical cross entropy, a typical choice for multi-class supervised classification tasks~\cite{IencoGDM17}.

Although the number of parameters of our architecture is not prohibitive, training of such models might be difficult and the final model can suffer by overfitting~\cite{DahlSH13}. With the aim to avoid such phenomena, following common practice for the training of Deep Learning architecture we leverage Dropout~\cite{DahlSH13} and Data Augmentation~\cite{abs-1712-04621}.

Dropout has been proposed to avoid co-adaptation of neurons during training~\cite{DahlSH13}. Dropout randomly “turning off” a given percentage of neurons (dropout rate hyperparameter) and their connections, corresponds to train a different, less correlated, model at every epoch. At inference time the neuron contribution is weighted by the dropout rate.
In our architecture we decide to apply Dropout (rate equals to 0.4) on the feature sets extracted by the two branches of \method just before the concatenation operation. This strategy will avoid to extract co-adapted features among the set of features successively employed to make the final decision.

Data augmentation~\cite{abs-1712-04621} is a common strategy to further increase the size of the training set and achieve higher model generalization. It consists in creating new synthetic training examples from those already available, by applying label preserving (random) transformations. In our case the (random) transformations are sampled from standard data augmentation techniques (90 degree rotation, vertical/horizontal flips and transpose). For each example, each technique is simultaneously performed on both the PAN and the corresponding MS patch. On average, the final training set has a size around three times the original training set.

\section{Data}
\label{sec:data}
\label{sec:data}

\subsection{REUNION dataset}

\begin{figure}[!ht]
\centering
\subfloat[\label{fig:cm_psR}] {\includegraphics[width=.48\linewidth]{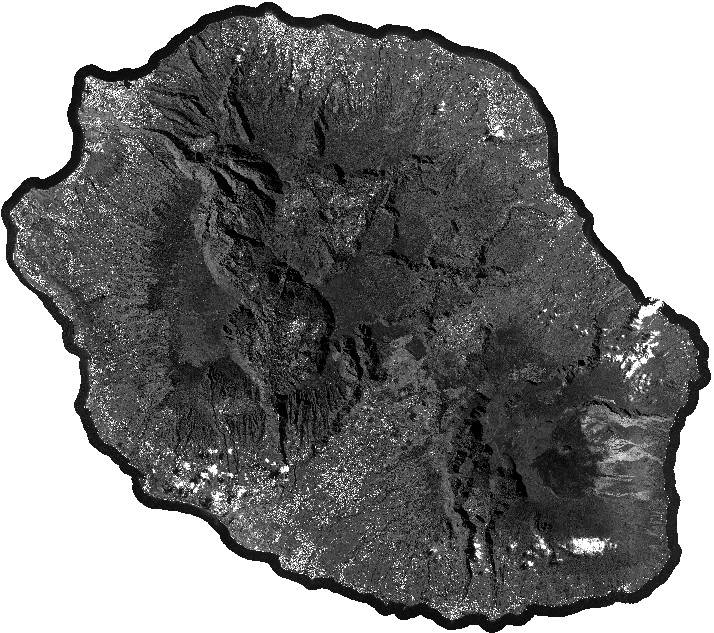} }
\subfloat[\label{fig:cm_dmilR}] {\includegraphics[width=.48\linewidth]{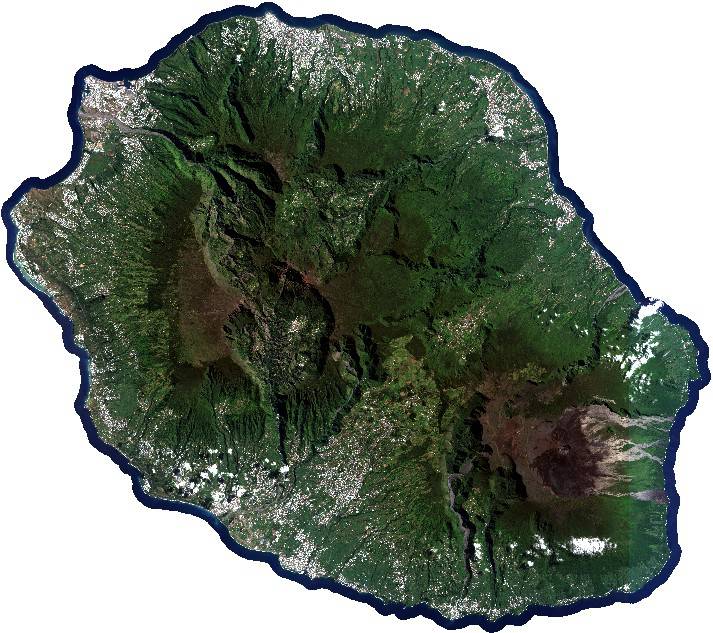}}
%\subfloat[\label{fig:cm_ourR}]{\includegraphics[width=.24\linewidth]{img_gard/gard_gt.png}}
\caption{ Panchromatic (a) and Multi Spectral (b) SPOT6 images of the \textit{REUNION} site. \label{fig:Reunion_PAN_XS} }
\end{figure}

We use a SPOT6 image, acquired on April 6th 2016 consisting of a 1.5~m Panchromatic band and 4 multi spectral bands (blue, green, red and near infrared) at 6~m resolution \textit{Top of Atmosphere} reflectance.
The Panchromatic image has a size of 44\,374 $\times$ 39\,422 while the Multi Spectral one has a size of 
11\,094 $\times$ 9\,856. %The image projection is Lambert93 (EPSG2154).% [ko] pas pour la réunion. 
Panchromatic and Multi Spectral satellite images are reported in Figure~\ref{fig:Reunion_PAN_XS}

%Its final size is 33\,280 $\times$ 29\,565 pixels on 5 bands (4 \textit{Top of Atmosphere} reflectance plus the NDVI). This image was also used as a reference to realign the different images in the time series by searching and mapping anchor points, in order to improve the spatial coherence between the different sources.

The field database, constituting the ground truth, was built from various sources: (i) the \textit{Registre parcellaire graphique} (RPG) \footnote{RPG is part of the European Land Parcel Identification System (LPIS), provided by the French Agency for services and payment} reference data of 2014, (ii) GPS records from June 2017 and (iii) photo interpretation of the VHSR image conducted by an expert, with knowledge of the territory, for distinguishing between natural and urban areas. All polygon contours have been resumed using the VHSR image as a reference. The final dataset is composed of a total of 464\,650 pixels  distributed over 13 classes, as indicated in Table~\ref{tab:data_reu}.

\begin{table}[!ht]
\centering
\begin{tabular}{|l||c|c|c|}
	\hline
\textbf{Class} & Label & \# \textbf{Objects} & \# \textbf{Pixels} \\ 
\hline \hline
1 & {\em Crop Cultivations} & 168 & 50061 \\ \hline
2 & {\em Sugar cane} & 167 & 50100  \\ \hline
3 & {\em Orchards} & 167 & 50092 \\ \hline
4 & {\em Forest plantations} & 67 & 20100 \\ \hline
5 & {\em Meadow} & 167 & 50100 \\ \hline
6 & {\em Forest} & 167 & 50100 \\ \hline
7 & {\em Shrubby savannah} & 173 & 50263 \\ \hline
8 & {\em Herbaceous savannah} & 78 & 23302 \\ \hline
9 & {\em Bare rocks} & 107 & 31587 \\ \hline
10 & {\em Urban areas} & 125 & 36046 \\ \hline
11 & {\em Greenhouse crops}& 49 & 14387 \\ \hline
12 & {\em Water Surfaces} & 96 &  2711\\ \hline
13 & {\em Shadows} & 38 & 11400 \\ \hline
\end{tabular}
\caption{Characteristics of the Reunion Dataset\label{tab:data_reu}}
\end{table}

%Total of 464650 pixels

\subsection{GARD dataset}

\begin{figure}[!ht]
\centering
\subfloat[\label{fig:cm_psR}] {\includegraphics[width=.48\linewidth]{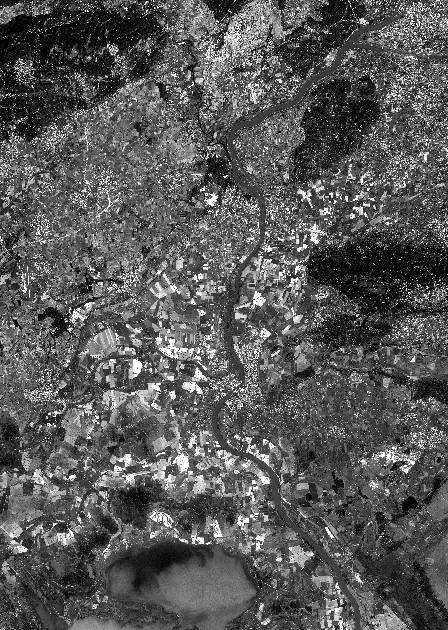} }
\subfloat[\label{fig:cm_dmilR}] {\includegraphics[width=.48\linewidth]{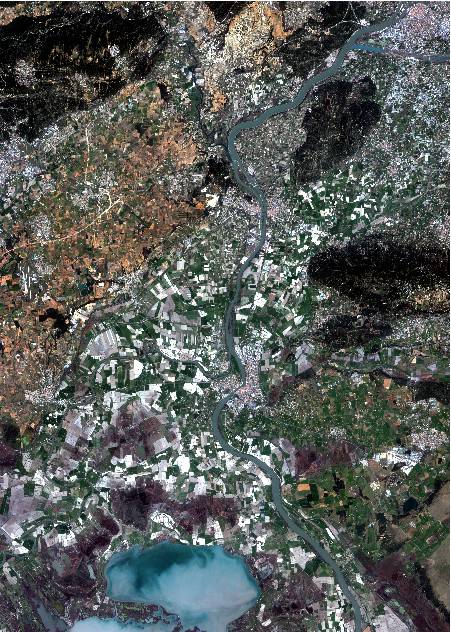}}
%\subfloat[\label{fig:cm_ourR}]{\includegraphics[width=.24\linewidth]{img_gard/gard_gt.png}}
\caption{ Panchromatic (a) and Multi Spectral (b) SPOT6 images of the \textit{GARD} site. \label{fig:Gard_PAN_XS} }
\end{figure}

The SPOT6 image, acquired on March 12th 2016 consists of a 1.5~m Panchromatic band and 4 Multi Spectral bands (Blue, Green, Red and Near Infrared) at 6~m resolution \textit{Top of Atmosphere} reflectance.
The Panchromatic image has a size of 24\,110 $\times$ 33\,740 while the Multi Spectral image has a size of 
6\,028 $\times$ 8\,435. Panchromatic and Multi Spectral satellite images are reported in Figure~\ref{fig:Gard_PAN_XS}.

%We used national scale thematic vector layers provided by IGN in order to mask irrelevant surfaces from satellite images. Buildings, transport infrastructures and water surfaces were extracted from the TOPO® Database. Closed forests were extracted from the Forest Database. 

%We also used the “Registre Parcellaire Graphique” (RPG) that is part of the European Land Parcel Identification System (LPIS), provided by the French Agency for services and payment (maintenant IGN). This is a land parcel identification and spatial information system for agricultural subsidies management. The RPG register represents the vector support for the annual declaration of land parcels under the Common Agricultural Policy (CAP) based on 20 classes. This information is particularly useful to define crop land cover, and it was used for training and validation sites in order to distinguish semi-natural areas from crops.

%The image projection is Lambert93 (EPSG2154).

The field database, constituting the ground truth, was built from various sources: (i) the \textit{Registre parcellaire graphique} (RPG)\footnotemark[\value{footnote}] reference data of 2016 and (ii) photo interpretation of the VHSR image conducted by an expert, with knowledge of the territory, for distinguishing between natural and urban areas. All polygon contours have been resumed using the VHSR image as a reference. The final dataset is composed of a total of 400\,970 pixels  distributed over 8 classes, as indicated in Table~\ref{tab:data_reu}.

\begin{table}[!ht]
\centering
\begin{tabular}{|l||c|c|c|}
	\hline
\textbf{Class} & Label & \# \textbf{Objects} & \# \textbf{Pixels} \\ 
\hline \hline
%Blé, orge, riz
1 & {\em Cereal Crops} & 167 & 50100 \\ \hline
%tournesol, pois_chiche
2 & {\em Other Crops} & 167 & 50098  \\ \hline
%verges, olivier
3 & {\em Tree Crops} & 167 & 50027 \\ \hline
%prairies
4 & {\em Meadows} & 167 & 49997 \\ \hline
%vignes
5 & {\em Vineyard} & 167 & 50100 \\ \hline
%foret
6 & {\em Forest} & 172 & 50273 \\ \hline
%bati
7 & {\em Urban areas} & 222 & 50275 \\ \hline
%eau
8 & {\em Water Surfaces} & 167 & 50100 \\ \hline
\end{tabular}
\caption{Characteristics of the Gard Dataset\label{tab:data_gard}}
\end{table}

\section{Experiments}
\label{sec:expe}
In this section, we present and discuss the experimental results obtained on the study sites introduced in Section~\ref{sec:data}. 

In order to provide a suitable insight on the behavior of \method, we perform different kinds of analysis. Firstly we analyze the global classification performances considering different evaluation metrics, secondly we inspect the per-class results of the different approaches and, finally, we supply a qualitative discussion considering the land cover maps produced by our framework.

\subsection{Competitors}
With the purpose to compare our approach (\method{}) to techniques tailored for the classification of Very High Spatial Resolution images we involve in the analysis three different competitors. The first two competitors take as input the Pansharpened image resulting from the combination between the Panchromatic and the Multi Spectral images. The Pansharpened image is obtained by the Bayesian Data Fusion technique~\cite{4505280}. For this purpose, we employ the public available implementation supplied by the Orfeo ToolBox\cite{Grizonnet2017}. The first competitor is a  Random Forest classifier (RF) that is trained on the set of radiometric values (R, G, B, NIR) associated to each pixel. We set the number of trees to 400. The second competitor is a Convolutional Neural Network classifier that has the same architecture structure of the $P$-$CNN$ module with 256, 512 and 1024 as number of filters for each layer respectively. The number of filters is augmented w.r.t. the $P$-$CNN$ architecture due to the amount of radiometric information (four bands) as input of such CNN. We refer to this competitor as $CNN_{PS}$. As third competitor, we adopt the method proposed in~\cite{LiuJZZZLYT18}, named $DMIL$, which also works directly on the multi-resolution dataset avoiding the pansharpening step. 
%The main differences between \method{} and $DMIL$ are two: i) $DMIL$ upscales the Multi Spectral image to the same resolution of the Panchromatic one with the objective to work on inputs at the same resolutions and ii) $DMIL$ manages the Multi Spectral information as a flat vector without considering the spatial correlation carried out by this source of data. 

Furthermore, similarly to what is proposed in~\cite{IencoGDM17}, we also investigate the possibility to use the different Deep Learning approaches as feature extractors to obtain a new data representation that successively feeds, as input, standard Machine Learning classifiers. To this end, we feed a Random Forest Classifier (with a number of trees equals to 400) with the features extracted by each of the different Deep Learning methods. To refer to this setting we use the notation $RF(\cdot)$. For instance, $RF(\method)$ indicates the Random Forest trained on the representation (features) learned by \method{}. With the objective to supply a more complete evaluation scenario, We also consider a baseline in which a Random Forest classifier is trained on the data patches extracted from the pansharpened image. We refer to this method as $RF(PATCH)$.

\subsection{Experimental Setting}
All the Deep Learning methods (including \method{}) are implemented using the Python Tensorflow library. 

During the learning phase, we use the Adam method~\cite{KingmaB14} to learn the model parameters with a learning rate equal to $2 \cdot 10^{-4}$. The training process is conducted over 250 epochs. The model that reaches the lowest value of the cost function (at training time) is used in the test phase.

The data are prepared as follows. We set the value of $d$ to 32. Due to the PAN and MS ratio on SPOT6 images (such ratio is equal to 4) and the fact that the MS image contains four raw bands, we have PAN (resp. MS) patches of size (32 x 32 x 1) (resp. (8 x 8 x 4) ). The values are normalized, per spectral band, in the interval $[0,1]$. Considering the $CNN_{PS}$ and $RF(PATCH)$ approaches, these methods take as input a patch of size (32 x 32 x 4) coming from the Pansharpened image. For each pair of patches (PAN and MS), the label information is associated to the pixel in position (16,16) of the PAN patch. Note that the use of odd-sized patches would have implied a more rigorous choice of the patch centers, however, it would also have prevented the coherent alignment of corresponding PAN and MS patches.

We divide the dataset into two parts, one for learning and the other one for testing the performance of the supervised classification methods. We used 30\% of the objects for the training phase while the remaining 70\% are employed for the test phase, in order to force a relative parsimony in the training stage with respect to the available reference data while ensuring a more robust validation. We impose that pixels of the same object belong exclusively to the training or to the test set to avoid spatial bias in the evaluation procedure~\cite{IngladaVATMR17}. 

Table~\ref{tab:TrainingTime} reports the training time of each deep learning method on a workstation with an Intel (R) Xeon (R) CPU E5-2667 v4@3.20Ghz with 256 GB of RAM and TITAN X GPU. The average learning time are reported in Table~\ref{tab:TrainingTime}. $CNN_{PS}$ is the approach that demands more time. $DMIL$ and \method{} consume very similar training time. The difference among the different methods is due to the fact that $CNN_{PS}$, for a fixed geographical area, needs to manage more information as input.

\begin{table}[!ht]
\centering
\scriptsize
\begin{tabular}{cccc}
Dataset & $CNN_{PS}$ & $DMIL$ & \method{} \\ \hline \hline
\textit{Gard} & 9h30m & 7h20m & 7h30m \\ \hline
\textit{Reunion} & 11h10m & 8h30m & 8h30m \\ \hline
\end{tabular}
\caption{Training time of the different Deep Learning approaches on the two study sites \label{tab:TrainingTime}}
\end{table}

The assessment of the classification performances is done considering global precision (\textit{Accuracy}), \textit{F-Measure}~\cite{IencoGDM17} and \textit{Kappa} measures.

It is known that, depending on the split of the data, the performances of the different methods may vary as simpler or more difficult examples are involved in the training or test set. To alleviate this issue, for each dataset and for each evaluation metric, we report results averaged over ten different random splits performed with the strategy previously presented.

\subsection{General Classification Results}
Table~\ref{tab:GardGen} and Table~\ref{tab:ReunionGen} summarize the results obtained by the different methods on the \textit{Gard} (resp. \textit{Reunion}) study site. The first (upper part) of each Table shows the performances of the main competing methods ($RF$, $CNN_{PS}$, $DMIL$ and \method{}) while the second (lower part) summarizes the results of the Random Forest classifier trained on the features learned by the different Deep Learning architectures and the baseline in which the input data involves patches of size (32 x 32 x 4) obtained from the Pansharpened image. 

Considering the results of the main competing approaches, we can observe that \method{} always outperforms the other approaches for all the three evaluation metrics. Systematically, \method{} obtains the best performances followed by $CNN_{PS}$, $DMIL$ and $RF$ respectively. We can also underline that, considering both study sites, the difference between the best and the second best result is always higher than four points for the \textit{Accuracy} measure. Similar behavior is exhibited considering the other evaluation metrics.
These experimental findings support our intuition that, when using a CNN-based deep learning approach for land cover classification, letting the architecture exploit sources at their native resolution (considering both spatial and spectral information) is more adequate than performing a prior pansharpening.

Regarding the use of the different Deep Learning approaches to extract features that are successively injected as input to standard Machine Learning method, we can note that this practice do not degrade the classification performances while, most of the time, it results in an improvement of the classification performances (w.r.t. the deep learning classification counterparts) of one or two points considering the whole set of evaluation metrics.
We can also note that the $RF$ classifier combined with a patch based input ($RF(PATCH)$) supplies results that are competitive w.r.t. $CNN_{PS}$ and $DMIL$. Also considering the $RF(PATCH)$ competitor, \method{} still provides better performances than this baseline approach. 

Generally, performance trends are similar between the two study sites. We can note that better results are achieved on the \textit{Reunion Island} study w.r.t. those obtained on the \textit{Gard}. This can be explained by the fact that, the SPOT6 image acquired on the \textit{Reunion Island} depicts this site during a period in which contrasts among the considered classes are more evident. More in detail, crops are easy to observe and highly distinguishable. This point positively influences the learning phase of all the competing methods. On the other hand, the image describing the \textit{Gard} site is acquired at the end of March when crops are not yet visible and most of the image is covered by bare soil. 

This evaluation highlights that data quality and data informativeness related to seasonal behaviors (considering the task at hand) are crucial issues that (positively or negatively) impact the construction of effective classification models for land cover mapping when agricultural classes are involved.

\begin{table}[!ht]
\centering
\scriptsize
 \begin{tabular}{| l || c | c | c |} \hline
& \textit{ Accuracy} & \textit{ F-Measure} &\textit{ Kappa} \\ \hline \hline
  RF(PIXEL) 	& 25.91 $\pm$ 0.16 & 25.52 $\pm$ 0.11 & 0.1532 $\pm$ 0.18 \\ \hline
 $CNN_{PS}$ 	&  66.14 $\pm$ 0.78 & 65.80$\pm$ 0.77 & 0.6131 $\pm$ 0.0089	\\ \hline
 $DMIL$ &  61.96 $\pm$ 1.00 & 61.76 $\pm$	1.01 &	0.5652 $\pm$ 0.0115 \\ \hline 
  \method & \textbf{70.48} $\pm$ 0.55 & \textbf{70.19} $\pm$ 0.67 & \textbf{0.6627} $\pm$ 0.0063	\\ 
 \hline \hline
 RF(PATCH)	& 69.93 $\pm$ 0.76 &  69.55 $\pm$ 0.77 & 0.6564 $\pm$ 0.87 \\ \hline 
 RF($CNN_{PS}$) & 68.04 $\pm$ 0.82 & 67.72 $\pm$ 0.84 & 0.6348 $\pm$ 0.0093 \\ \hline
RF(DMIL) & 64.79 $\pm$ 0.73 & 64.43 $\pm$ 0.82 & 0.5976 $\pm$
0.0084 \\ \hline
RF(\method) & \textbf{71.98} $\pm$ 0.58 & \textbf{71.73} $\pm$ 0.54 &	\textbf{0.6797} $\pm$ 0.0066 \\ 
\hline
\hline
 \end{tabular}
 \caption{GARD \label{tab:GardGen}}
\end{table}

\begin{table}[!ht]
\centering
\scriptsize
 \begin{tabular}{| l || c | c | c |} \hline
& \textit{ Accuracy} & \textit{ F-Measure} &\textit{ Kappa} \\ \hline \hline
 
 RF(PIXEL) 	& 24.87 $\pm$ 0.2  & 23.66 $\pm$ 0.2 & 0.1719 $\pm$ 0.0024	\\ \hline
 $CNN_{PS}$ & 74.49 $\pm$ 1.20 &  74.25 $\pm$ 1.24 & 0.7195 $\pm$ 0.0131 \\ \hline
 DMIL & 69.40 $\pm$ 1.11 & 69.34 $\pm$ 1.12	& 0.6637 $\pm$ 0.0121	\\ \hline 
  \method & \textbf{79.65} $\pm$ 0.87 & \textbf{79.56} $\pm$ 0.91 & \textbf{0.7764} $\pm$ 0.0096\\ \hline \hline
 $RF(PATCH)$	& 72.22 $\pm$ 1.31 & 71.53 $\pm$ 1.4 & 0.6943 $\pm$ 0.0144 \\ \hline 
$RF(CNN_{PS})$ & 75.77 $\pm$ 1.14 & 75.56 $\pm$ 1.19 & 0.7334 $\pm$ 0.0125\\ \hline
$RF(DMIL)$ & 71.98 $\pm$ 0.46 & 71.94 $\pm$ 0.47 & 0.6918 $\pm$ 0.51\\ \hline
$RF(\method)$ & \textbf{79.67} $\pm$ 0.82 & \textbf{79.52} $\pm$ 0.86 & \textbf{0.7763} $\pm$ 0.0090 \\ \hline

\hline
 \end{tabular}
 \caption{REUNION \label{tab:ReunionGen}}
\end{table}

\subsection{Per Class Classification Results}
Table~\ref{tab:PerClass_fm_reunion} and Table~\ref{tab:PerClass_fm_gard} depict the per class F-Measure results for the \textit{Reunion Island} and the \textit{Gard} study sites respectively. Also in this case, for each study site, we differentiate between the main competing methods ($RF$, $CNN_{PS}$, $DMIL$ and \method) and experiments in which, for each Deep Learning method, features are extracted and a Random Forest classifier is built upon these new representations. 

\begin{table*}[!ht]
\scriptsize
\centering
\begin{tabular}{|l||c|c|c|c|c|c|c|c|c|c|c|c|c|}
	\hline
\textbf{Method} & 1 & 2 & 3 & 4 & 5 & 6 & 7 & 8 & 9 & 10 & 11 & 12 & 13\\  \hline \hline
RF & 17.04\% & 24.3\% & 19.81\% & 30.58\% & 11.16\% & 31.99\% & 22.11\% & 10.29\% & 28.68\% & 49.15\% & 3.37\% & 11.75\% & 67.06\% \\ \hline
$CNN_{PS}$ & 64.49\% & 75.03\% & 69.38\% & \textbf{83.97}\% & 67.63\% & 77.69\% & 74.63\% & 57.47\% & 75.52\% & 83.06\% & 67.37\% & 93.04\% & \textbf{96.96}\% \\ \hline
$DMIL$ & 58.95\% & 69.13\% & 56.86\% & 81.63\% & 60.59\% & 72.37\% & 72.54\% & 56.13\% & 72.86\% & 79.67\% & 64.27\% & 92.56\% & 95.09\% \\ \hline
%\method & 70.92\% & 77.53\% & 77.12\% & 85.87\% & 71.97\% & 81.85\% & 79.15\% & 63.2\% & 78.27\% & 86.68\% & 78.56\% & 94.15\% & 95.19\% \\ \hline 
\method & \textbf{72.15}\% & \textbf{78.48}\% & \textbf{77.81}\% & 83.82\% & \textbf{73.97}\% & \textbf{81.52}\% & \textbf{79.48}\% & \textbf{66.58}\% & \textbf{79.69}\% & \textbf{87.91}\% & \textbf{80.23}\% & \textbf{95.23}\% & 94.78\% \\ \hline \hline

RF(PATCH) & 59.47\% & 75.02\% & 67.89\% & 81.99\% & 67.67\% & 75.28\% & 73.33\% & \textbf{63.46}\% & 73.62\% & 75.15\% & 31.95\% & 93.01\% & \textbf{97.93}\% \\ \hline
RF($CNN_{PS}$) & 66.18\% & 77.06\% & 71.44\% & 84.55\% & 69.25\% & 79.98\% & 76.11\% & 59.36\% & 75.51\% & 83.04\% & 65.9\% & 93.38\% & 97.69\% \\ \hline
RF($DMIL$) & 60.8\% & 72.21\% & 61.14\% & 83.84\% & 65.56\% & 75.41\% & 75.08\% & 58.76\% & 74.25\% & 80.84\% & 65.03\% & 93.07\% & 95.37\% \\ \hline
%RF(\method) & 72.0\% & 78.59\% & 77.2\% & 86.06\% & 73.47\% & 83.19\% & 79.58\% & 62.82\% & 80.81\% & 86.16\% & 74.86\% & 95.28\% & 96.59 \\ \hline
RF(\method) & \textbf{71.33}\% & \textbf{79.05}\% & \textbf{77.94}\% & \textbf{86.04}\% & \textbf{74.57}\% & \textbf{83.31}\% & \textbf{79.65}\% & 63.42\% & \textbf{80.71}\% & \textbf{86.15}\% & \textbf{73.81}\% & \textbf{94.89}\% & 96.88\% \\ \hline
\end{tabular}
\caption{Reunion Dataset\label{tab:PerClass_fm_reunion}}
\end{table*}

\begin{table*}[!ht]
\centering
\begin{tabular}{|l||c|c|c|c|c|c|c|c|}
	\hline
\textbf{Method} & 1 & 2 & 3 & 4 & 5 & 6 & 7 & 8 \\   \hline \hline
RF & 16.26\% & 18.86\% & 17.77\% & 17.62\% & 15.2\% & 52.2\% & 30.09\% & 36.06\% \\ \hline
$CNN_{PS}$ & 44.67\% & 57.04\% & 50.63\% & 51.0\% & 57.14\% & 86.75\% & 83.37\% & 95.59\% \\ \hline
$DMIL$ & 40.78\% & 51.1\% & 47.87\% & 47.63\% & 54.85\% & 86.16\% & 70.01\% & 95.52\% \\ \hline
%\method & 49.52\% & 61.12\% & 59.99\% & 56.39\% & 59.81\% & 89.31\% & 88.07\% & 95.64\% \\ \hline
\method & \textbf{51.74}\% & \textbf{60.2}\% & \textbf{58.51}\% & \textbf{55.36}\% & \textbf{61.15}\% & \textbf{89.07}\% & \textbf{88.94}\% & \textbf{96.3} \% \\ \hline
\hline \hline

RF(PATCH) & 48.28\% & 62.56\% & 56.54\% & 56.17\% & 62.13\% & 87.71\% & 85.25\% & \textbf{97.46}\% \\ \hline
RF($CNN_{PS}$) & 45.86\% & 60.07\% & 54.9\% & 53.56\% & 58.33\% & 88.33\% & 84.47\% & 96.06\% \\ \hline
RF($DMIL$) & 41.17\% & 54.89\% & 52.02\% & 51.83\% & 59.14\% & 86.97\% & 73.52\% & 95.76\% \\ \hline
%RF(\method) & 52.19\% & 64.46\% & 61.0\% & 58.68\% & 63.8\% & 89.3\% & 89.48\% & 96.47\% \\ \hline
RF(\method) & \textbf{52.53}\% & \textbf{64.13}\% & \textbf{60.49}\% & \textbf{58.14}\% & \textbf{63.85}\% & \textbf{89.1}\% & \textbf{88.95}\% & 96.46\% \\ \hline
\hline
\end{tabular}
\caption{Gard Dataset\label{tab:PerClass_fm_gard}}
\end{table*}

Considering the main competing approaches, we can observe that, for both study sites, \method obtains better or very similar per class \textit{F-Measure} w.r.t. the others approaches. For the classification of the \textit{Reunion Island} dataset, we can note significant improvement in classes (1),(3),(5),(8) and (11) (resp. \textit{Crop Cultivations}, \textit{Orchards}, \textit{Meadow}, \textit{Herbaceous savannah} and \textit{Greenhouse crops}). Here, the improvement ranges between six points (\textit{Meadow}) and twelve points (\textit{Greenhouse crops}) w.r.t. the second best method. When we analyze the per-class results on the \textit{Gard} site, we can note that also in this case we have clear amelioration for certain classes: (1) and (3) (\textit{Cereal Crops} and \textit{Tree Crops}) with an average gain of 7 points of \textit{Accuracy}.

Similarly to previous results, also in the per class analysis we can note that the Random Forest approach coupled with the features learned by the different methods (lower part of Table~\ref{tab:PerClass_fm_reunion} and Table~\ref{tab:PerClass_fm_gard}) provides systematic improvement w.r.t. almost all the land cover classes compared to the pure deep learning classification approaches.

\begin{figure*}[!ht]
\centering
\subfloat[\label{fig:cm_psR}] {\includegraphics[width=.32\linewidth]{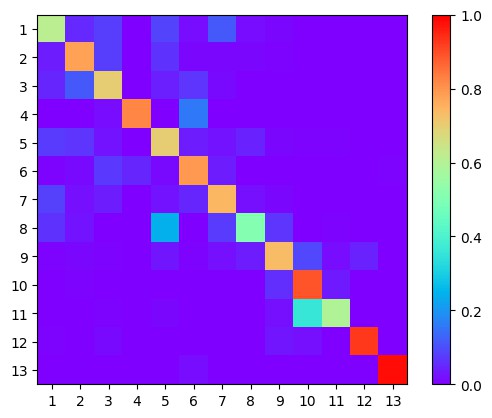} }
\subfloat[\label{fig:cm_dmilR}] {\includegraphics[width=.32\linewidth]{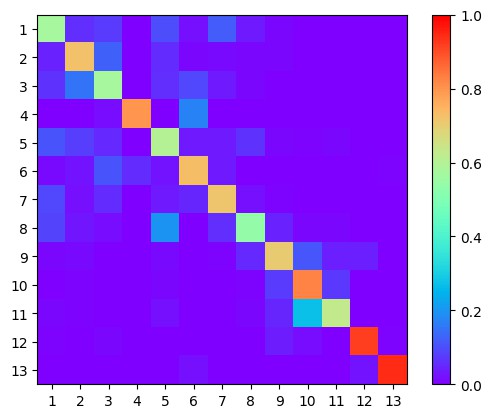}}
\subfloat[\label{fig:cm_ourR}]{\includegraphics[width=.32\linewidth]{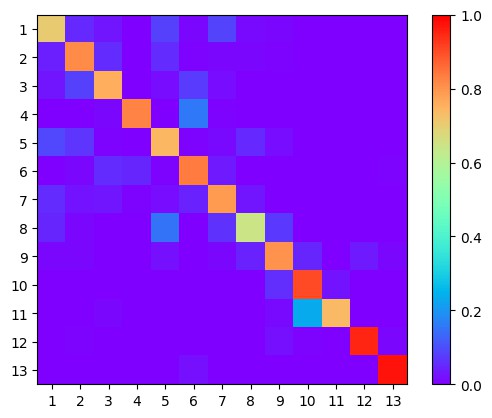}}
\newline
\subfloat[\label{fig:cm_psG}]{\includegraphics[width=.32\linewidth]{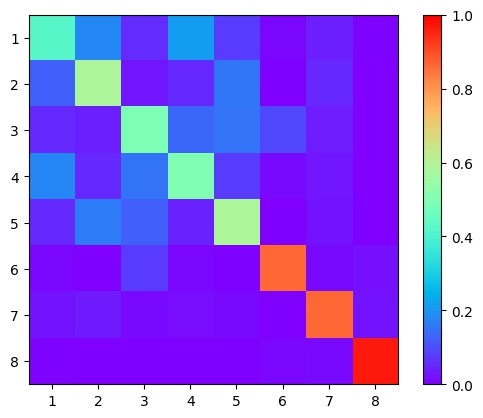}}
\subfloat[\label{fig:cm_dmilG}]
{\includegraphics[width=.32\linewidth]{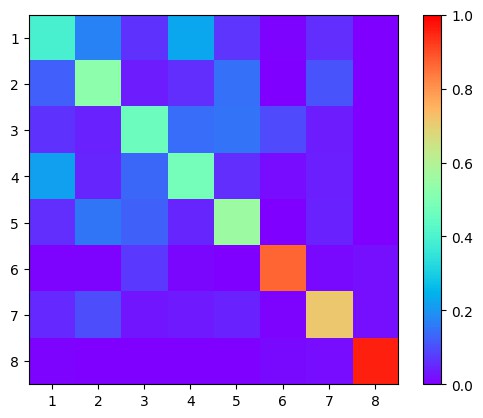} }
\subfloat[\label{fig:cm_ourG}]
{\includegraphics[width=.32\linewidth]{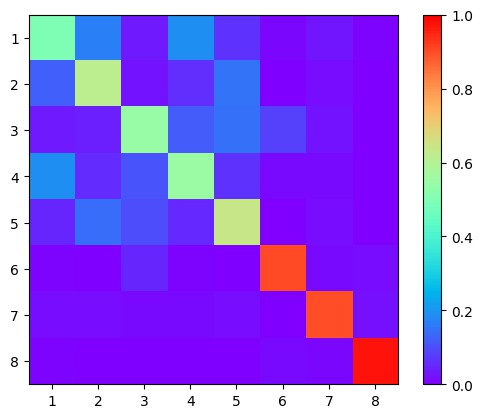}}
\caption{Confusion matrices of the Deep Learning approaches on the \textit{Reunion Island} dataset ( $CNN_{PS}$ (a),  $DMIL$ (b) and \method{} (c) ) and on the \textit{Gard} dataset ( $CNN_{PS}$ (d), $DMIL$ (e) and \method{} (f) ).}
\label{fig:confMatrix}
\end{figure*}

To further advance the understanding of our method, we report in Figure~\ref{fig:confMatrix} the confusion matrices associated to the $CNN_{PS}$, $DMIL$ and \method{} methods respectively on the two study sites.
Figure~\ref{fig:cm_psR}, Figure~\ref{fig:cm_dmilR} and Figure~\ref{fig:cm_ourR} depict the confusion matrices of $CNN_{PS}$, $DMIL$ and \method{} on the \textit{Reunion Island} study site. We can note that the confusion matrix associated to \method{} has clearly a stronger diagonal (towards dark red) compared to the confusion matrices of the other approaches. %This point is also highlighted by the fact that the confusion matrices in Figure~\ref{fig:cm_psR} and Figure~\ref{fig:cm_dmilR} ($CNN_{PS}$ and $DMIL$ resp.) have a more pixelized behavior outside the main diagonal.

Figure~\ref{fig:cm_psG}, Figure~\ref{fig:cm_dmilG} and Figure~\ref{fig:cm_ourG} represent the confusion matrices of $CNN_{PS}$, $DMIL$ and \method{}, respectively, on the \textit{Gard} study site. Here, the different confusion matrices share a more similar appearance w.r.t. those shown for the \textit{Reunion Island} dataset. Nevertheless, we can still observe a slightly more suitable behavior exhibited by \method: i) a slightly darker diagonal on both strong and weaker classes and ii) a generally less intense ``noise'' outside the diagonal compared to the competitors.

\subsection{Qualitative Inspection of Land Cover Map}
In Table~\ref{tab:gard_examples} and~\ref{tab:reunion_examples} we report some representative map classification details on the \textit{Gard} and \textit{Reunion Island} datasets considering the $DMIL$, $CNN_{PS}$ and \method, respectively.

\begin{table*}[!ht]
\centering
\begin{tabular}{cccc}
\textit{VHSR Image} & \textbf{$DMIL$} & \textbf{$CNN_{PS}$} & \textbf{ \method } \\ 
\subfloat[\label{fig:gard_ex01_pan}] {\includegraphics[width=.22\linewidth]{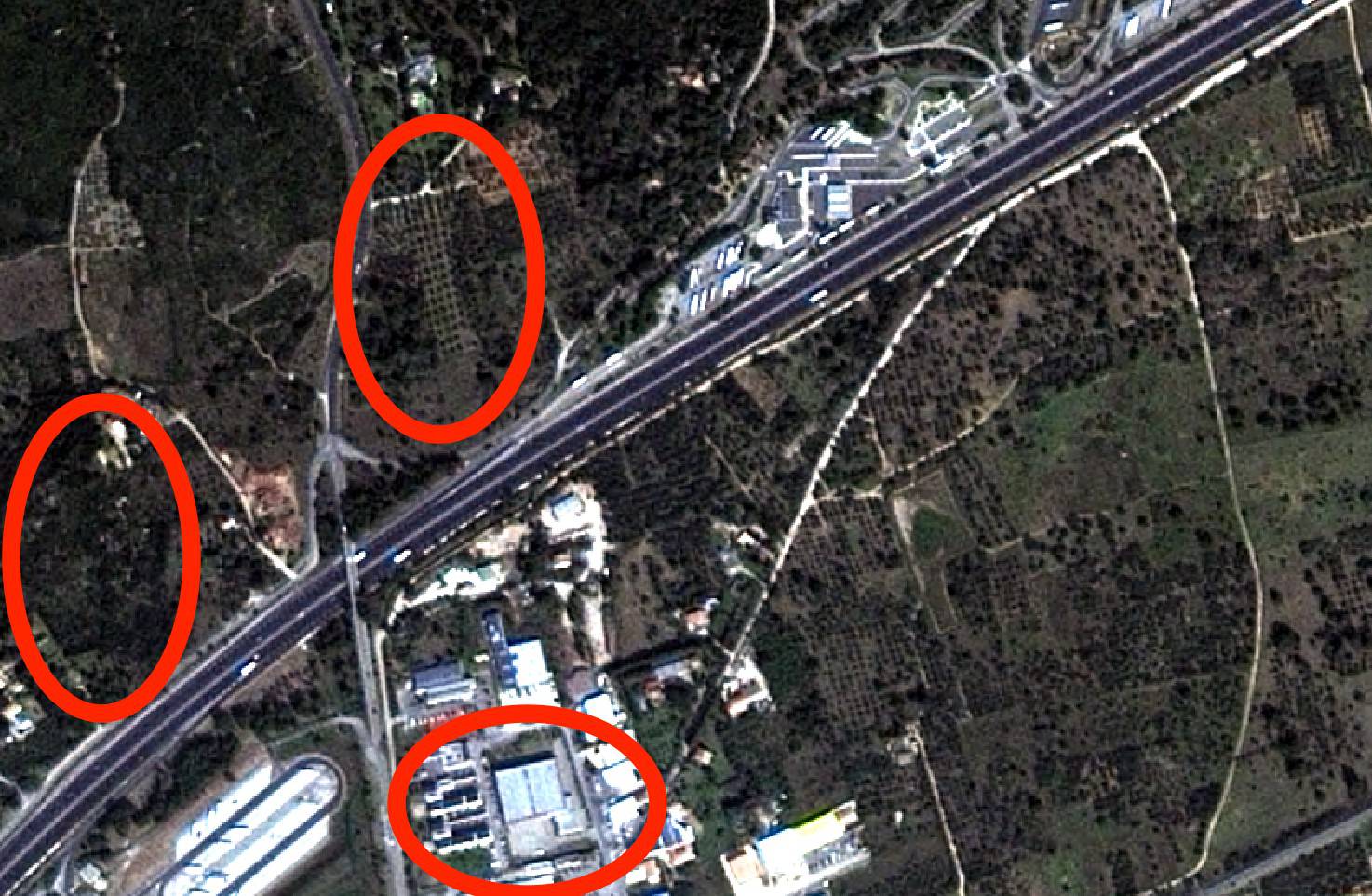} }
&
\subfloat[\label{fig:gard_ex01_dmil}] {\includegraphics[width=.22\linewidth]{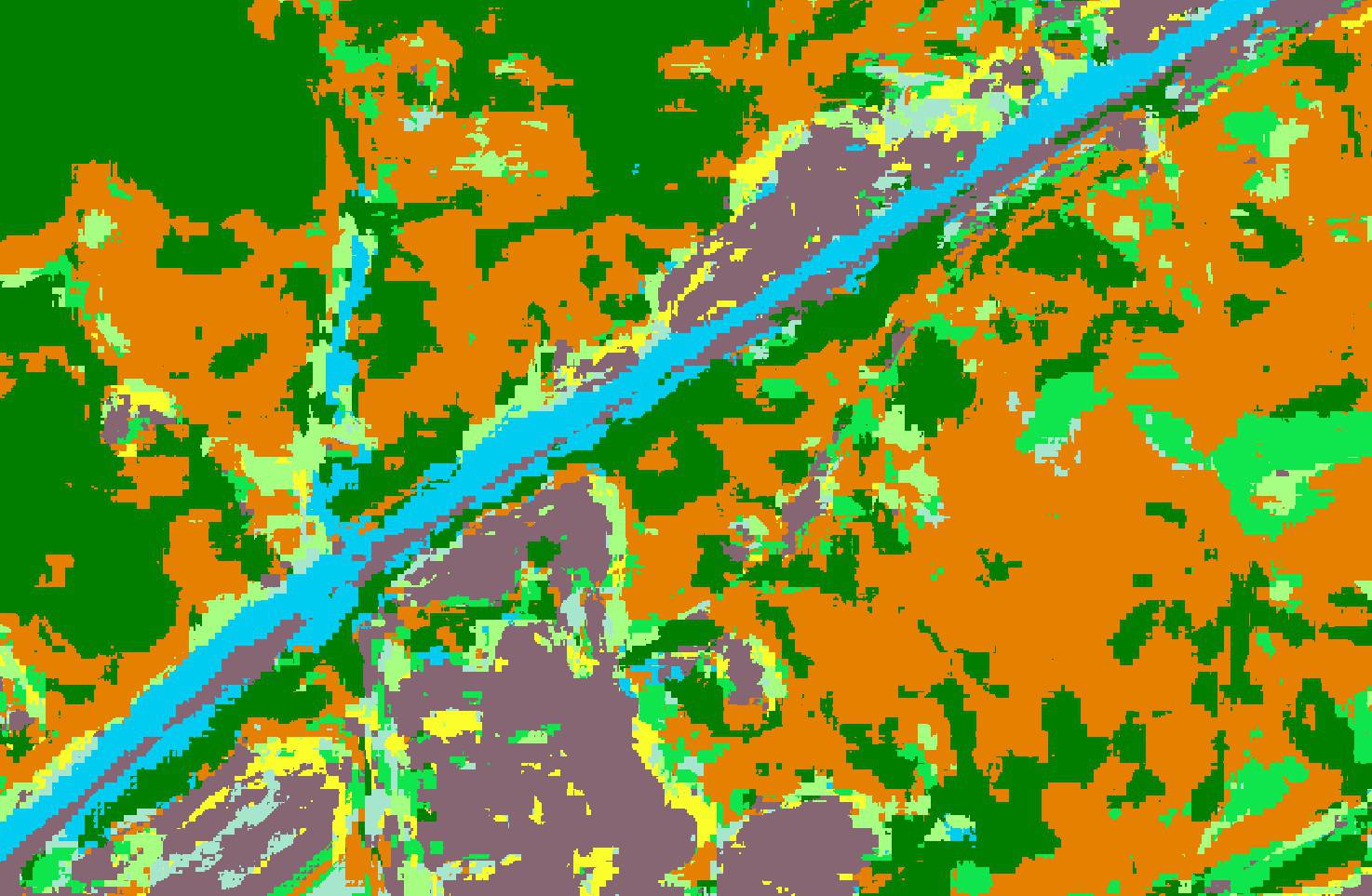} }
&
\subfloat[\label{fig:gard_ex01_ps}] {\includegraphics[width=.22\linewidth]{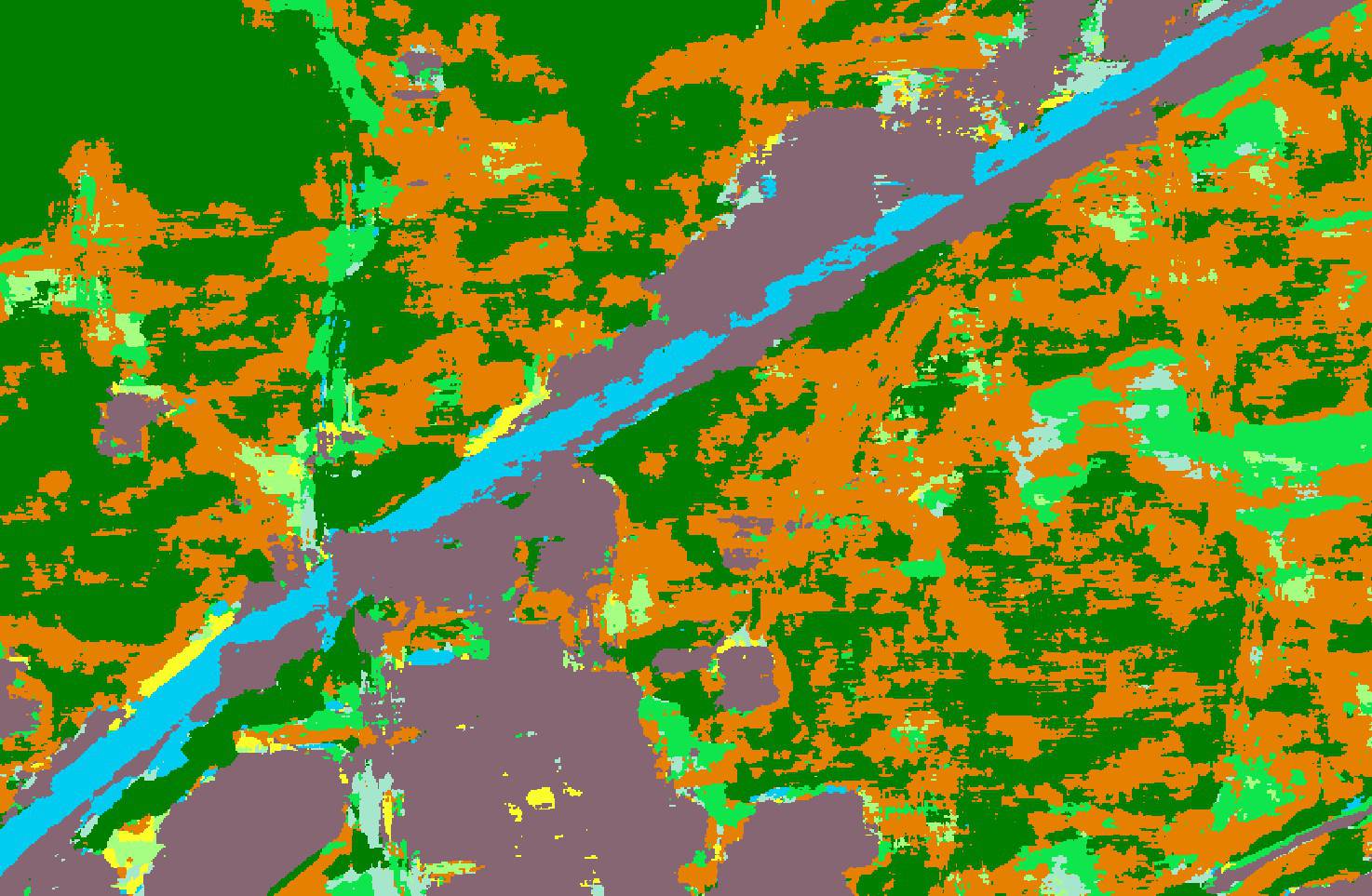} }
&
\subfloat[\label{fig:gard_ex01_our}] {\includegraphics[width=.22\linewidth]{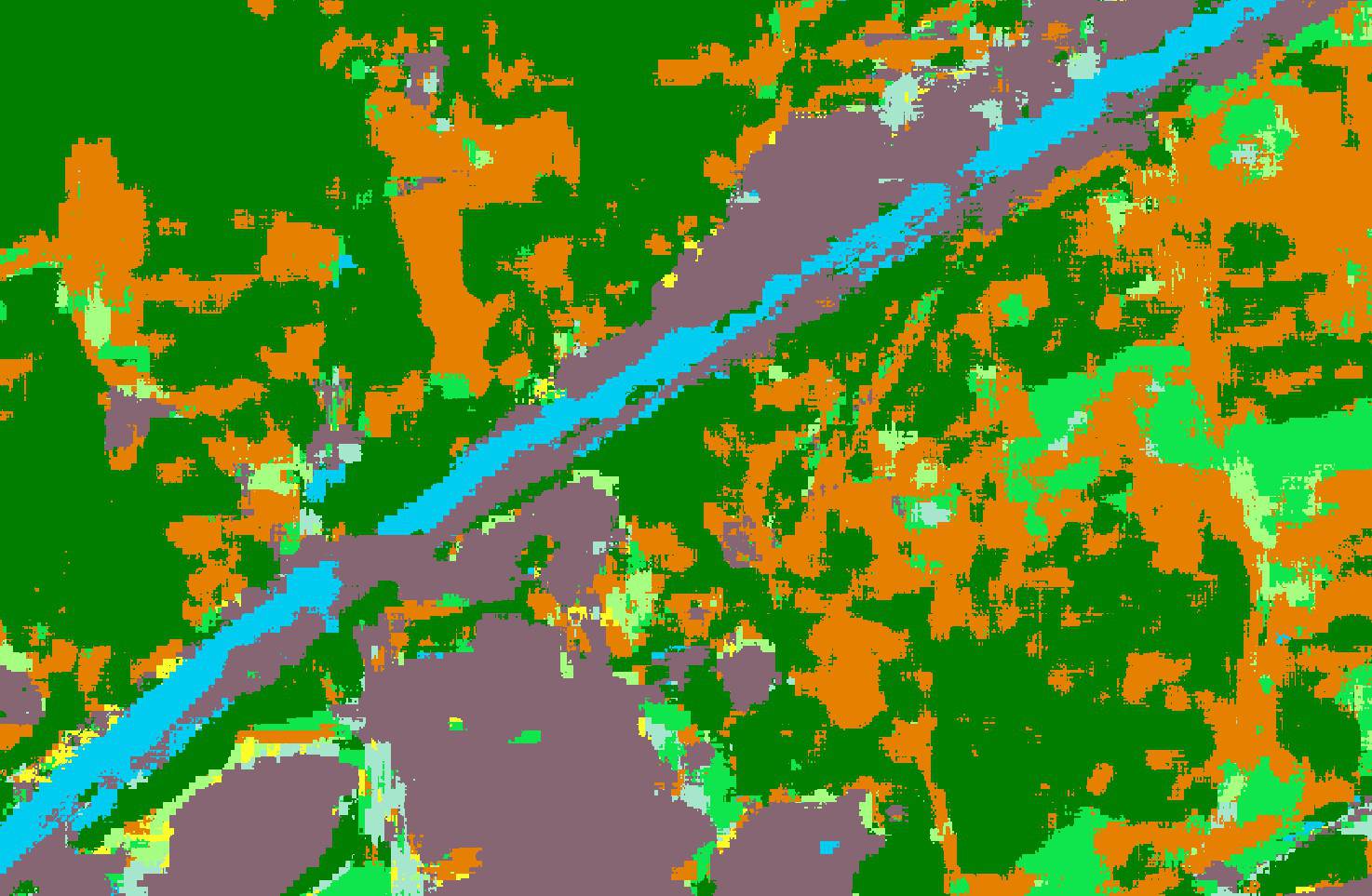} }
\\  
\subfloat[\label{fig:gard_ex02_pan}] {\includegraphics[width=.22\linewidth]{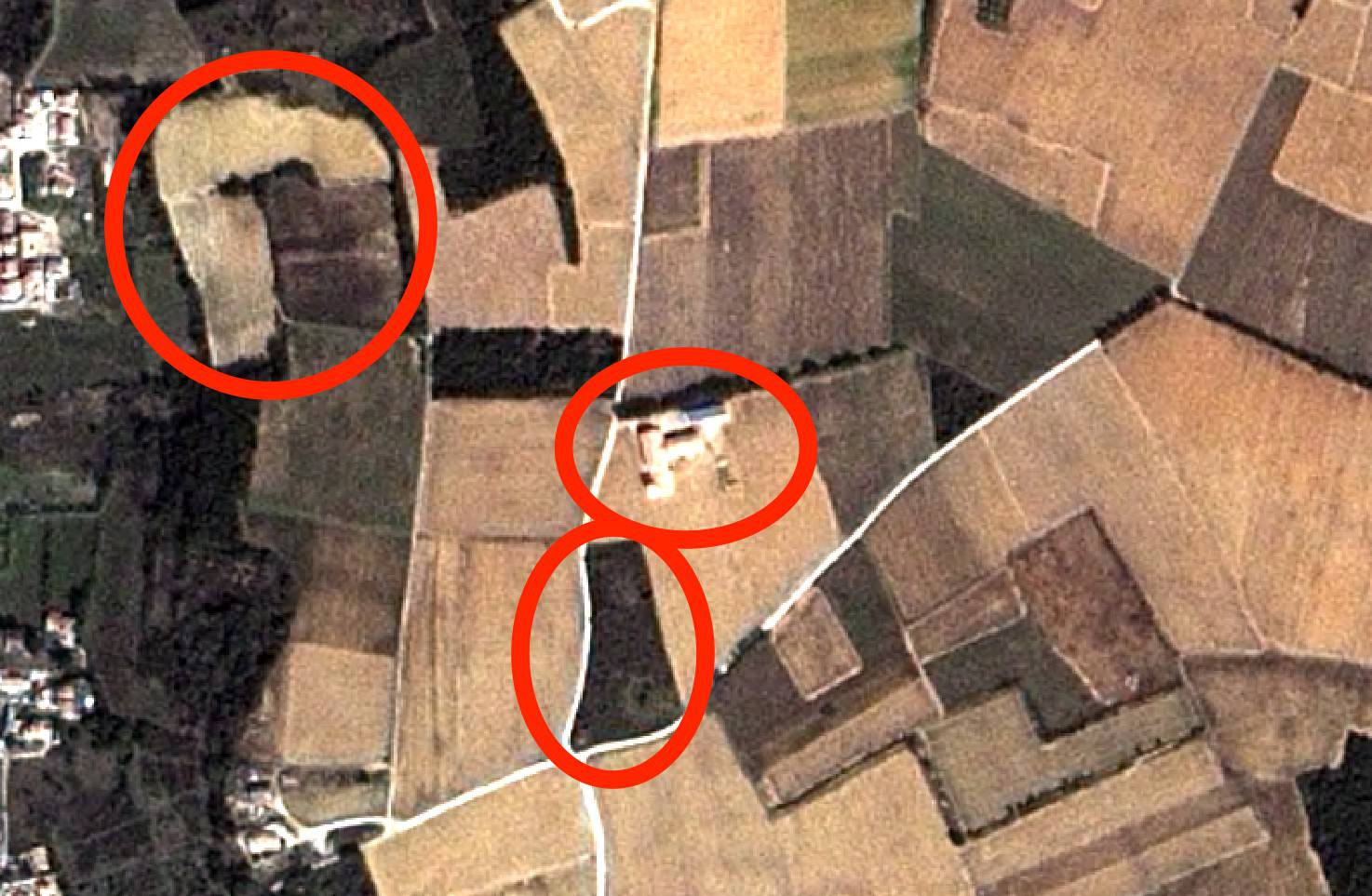} }
&
\subfloat[\label{fig:gard_ex02_dmil}] {\includegraphics[width=.22\linewidth]{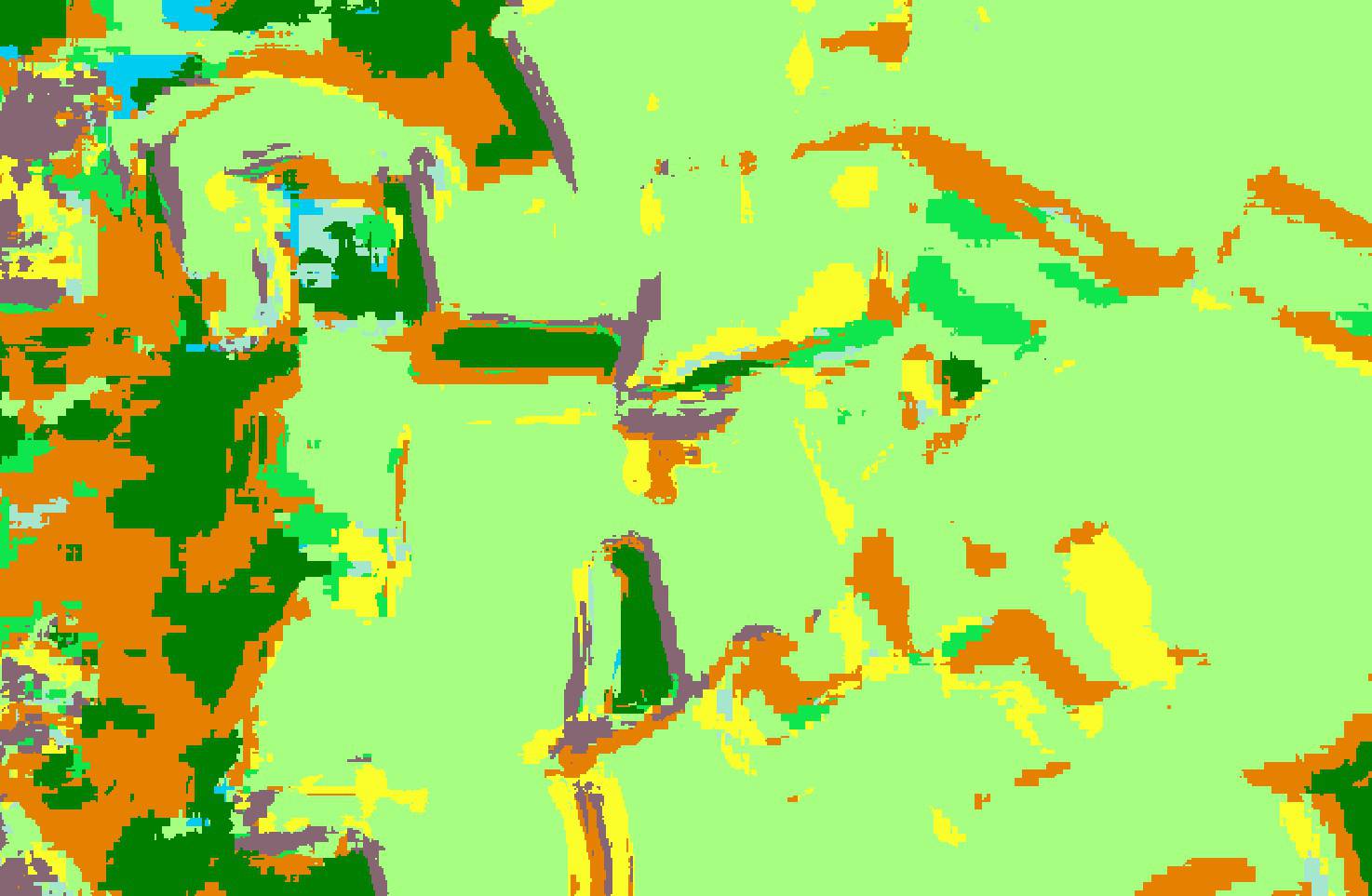} }
&
\subfloat[\label{fig:gard_ex02_ps}] {\includegraphics[width=.22\linewidth]{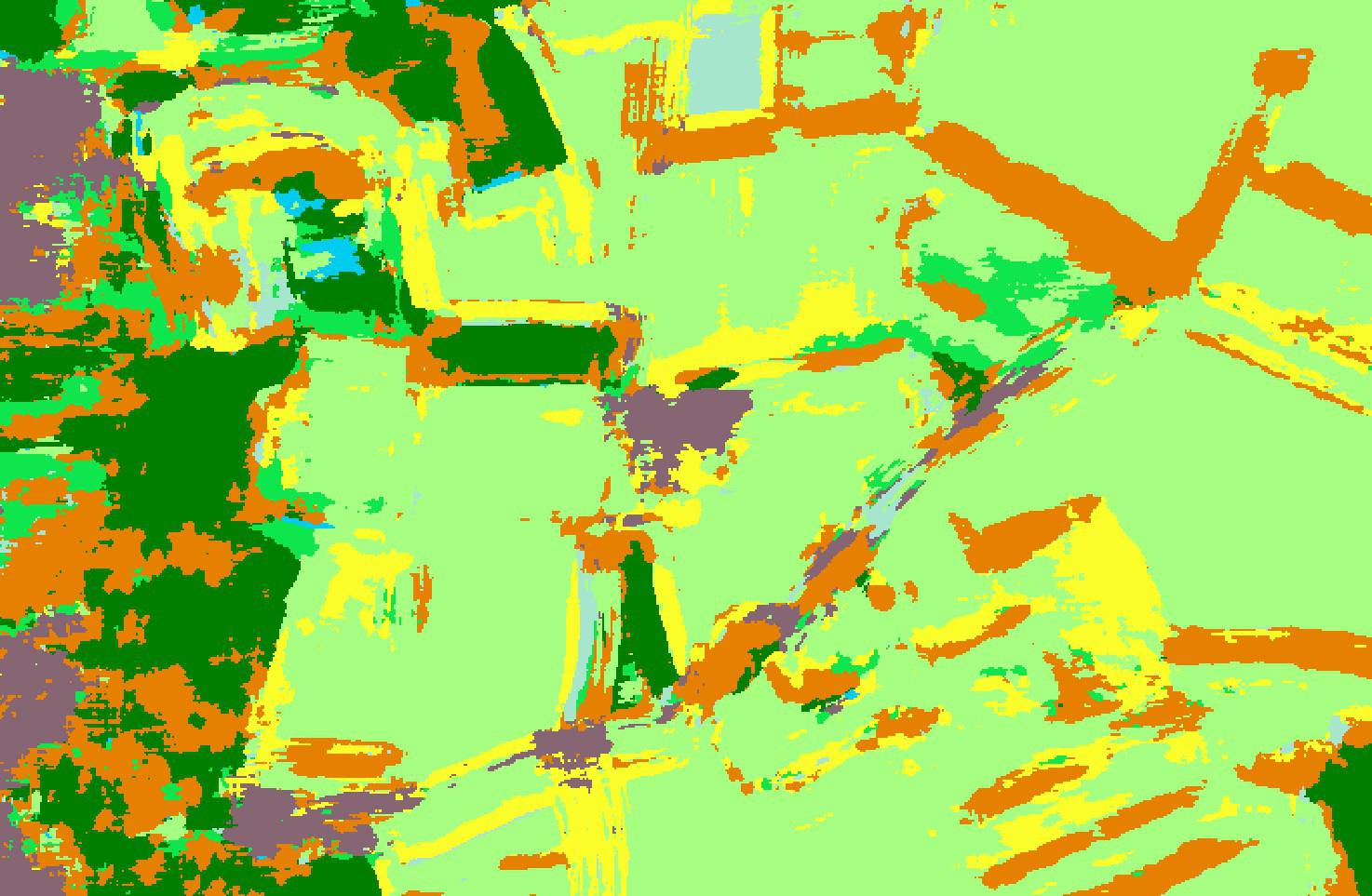} }
&
\subfloat[\label{fig:gard_ex02_our}] {\includegraphics[width=.22\linewidth]{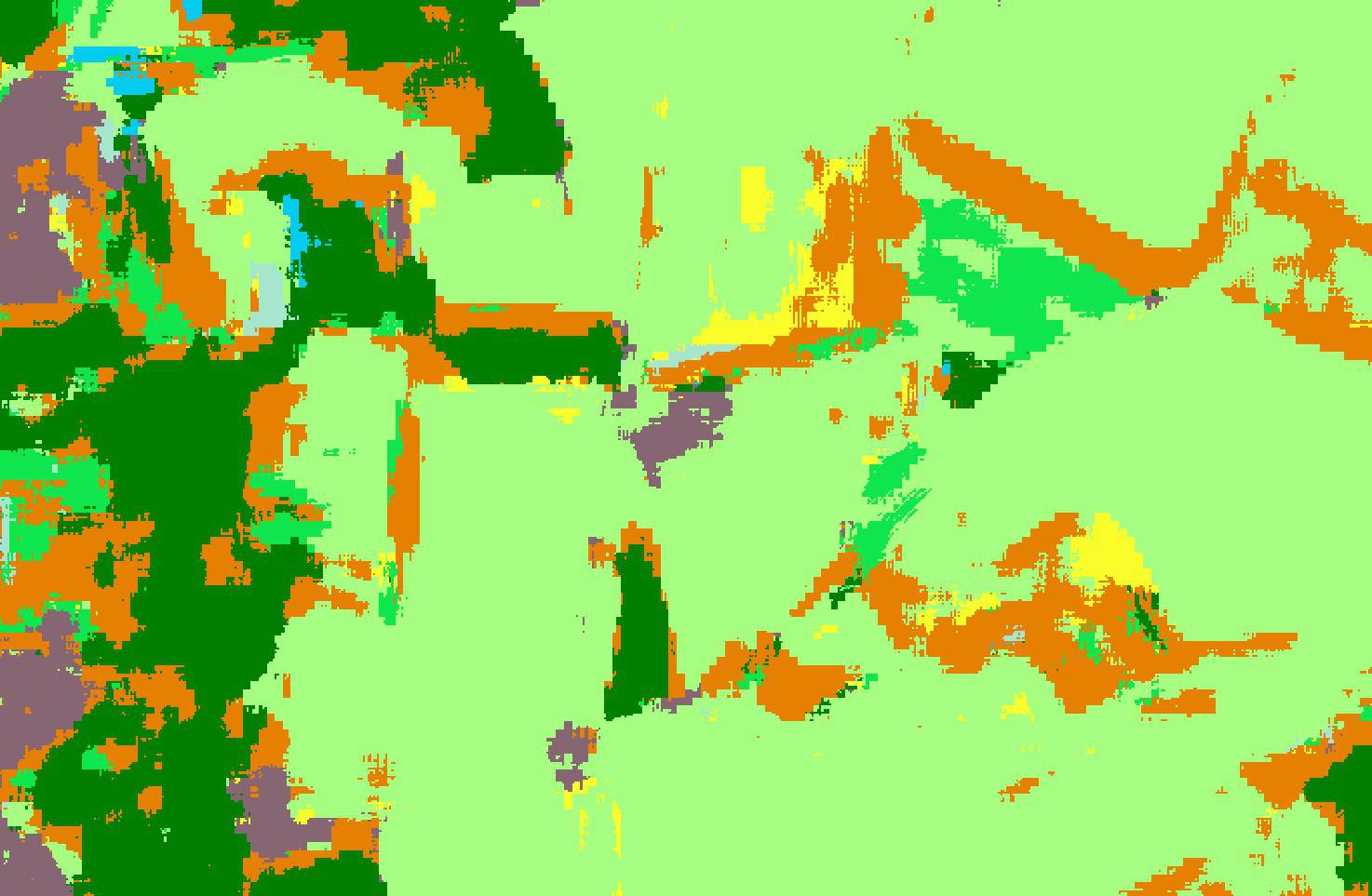} }

\\
\subfloat[\label{fig:gard_ex03_pan}] {\includegraphics[width=.22\linewidth]{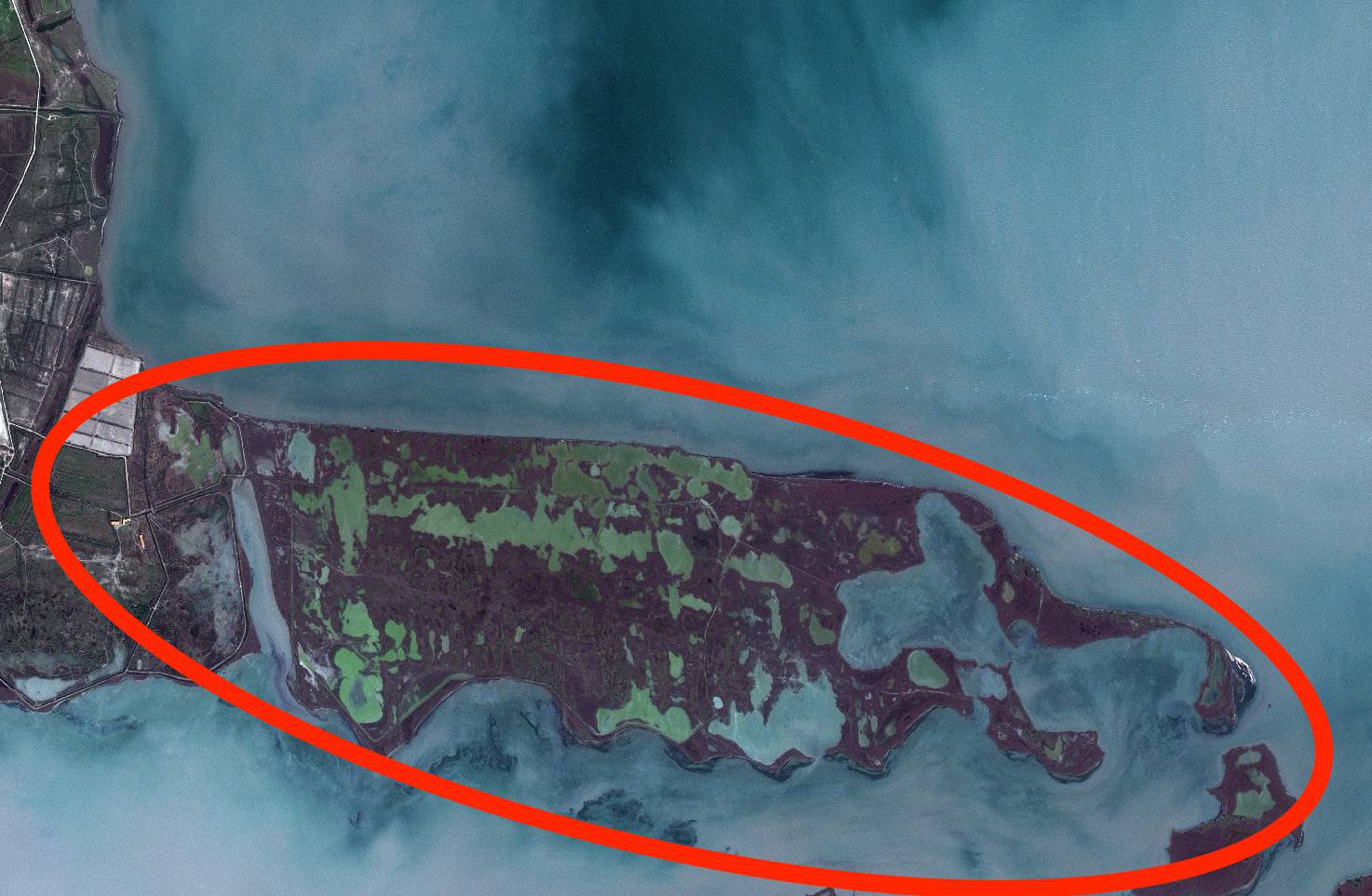} }
&
\subfloat[\label{fig:gard_ex03_dmil}] {\includegraphics[width=.22\linewidth]{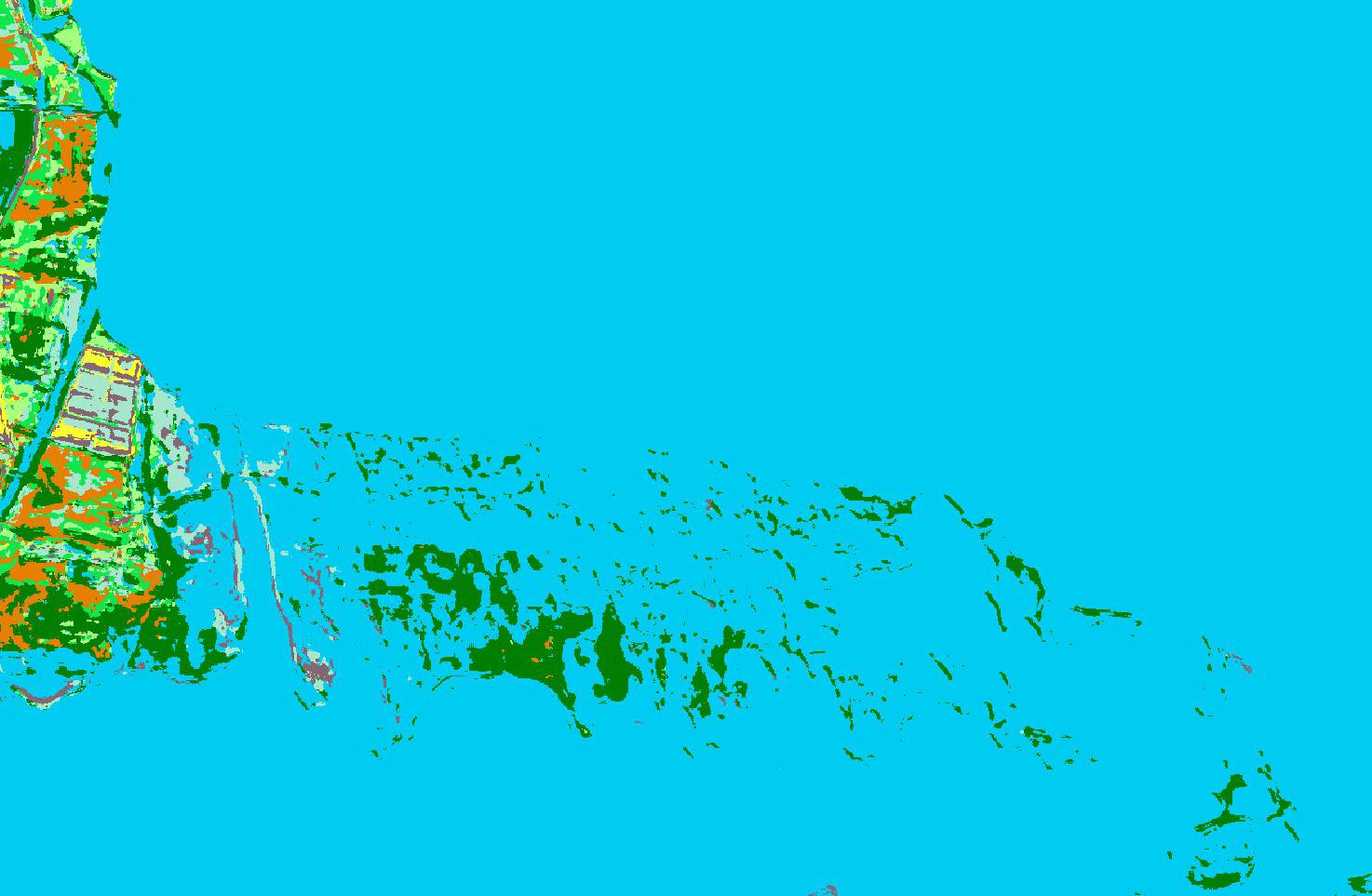} }
&
\subfloat[\label{fig:gard_ex03_ps}] {\includegraphics[width=.22\linewidth]{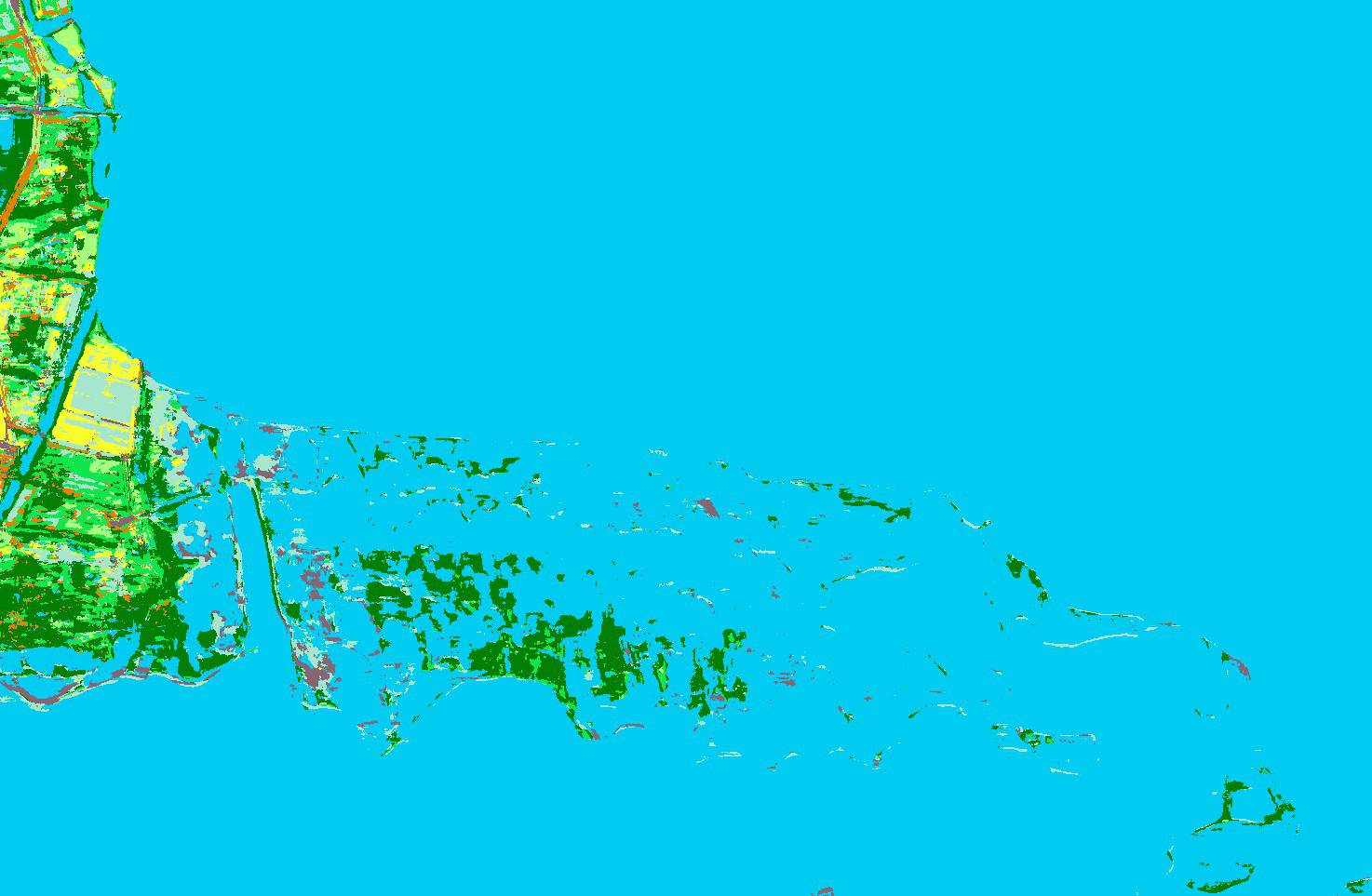} }
&
\subfloat[\label{fig:gard_ex03_our}] {\includegraphics[width=.22\linewidth]{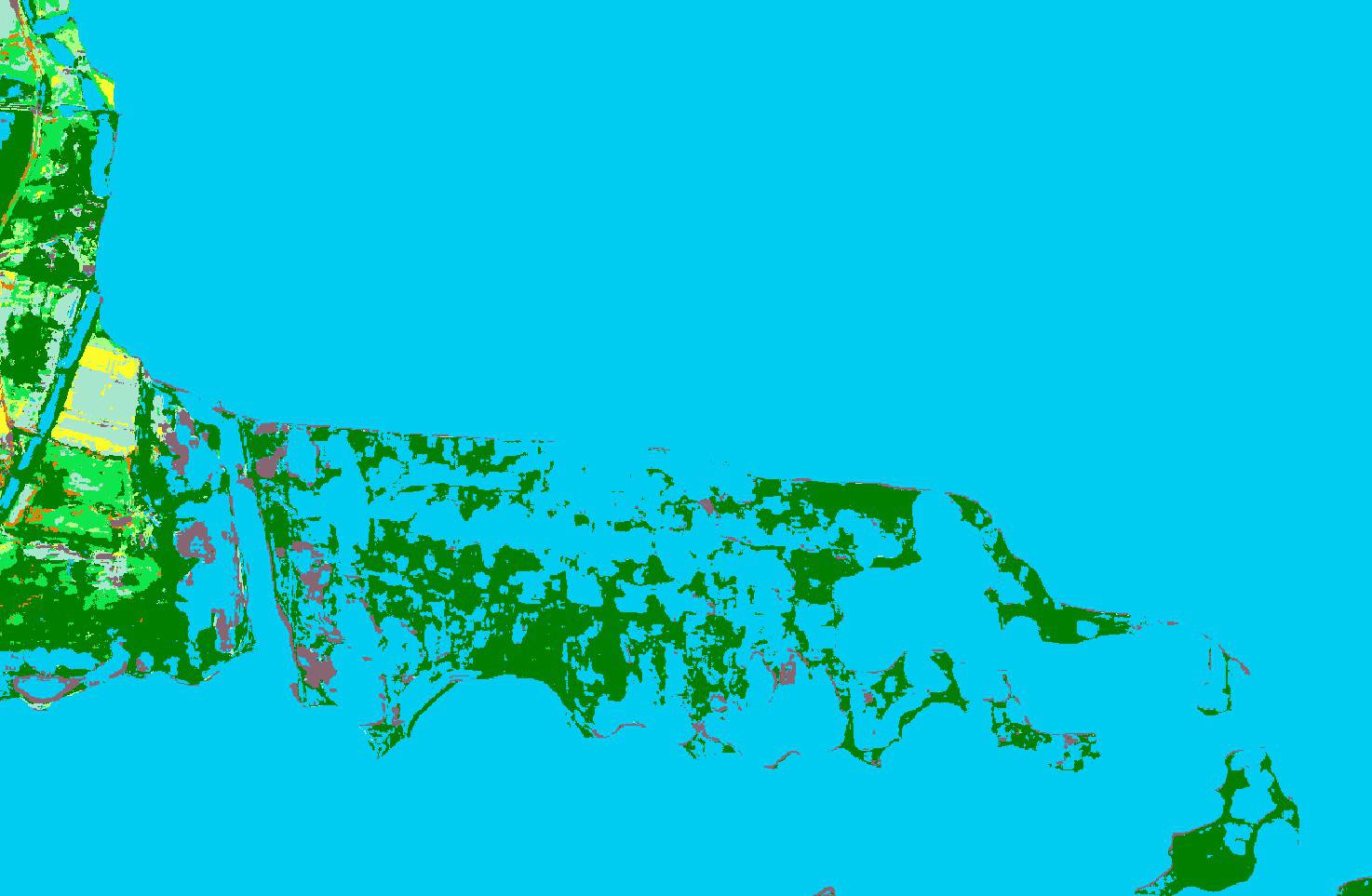} }
\\ 
% \\
\multicolumn{4}{c}{ \includegraphics[width=1.\textwidth]{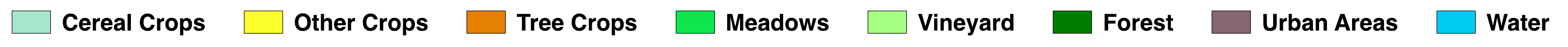}  }

\end{tabular}
\caption{table caption GARD \label{tab:gard_examples}}
\end{table*}

Regarding the \textit{Gard} study site, the first example (Figures~\ref{fig:gard_ex01_pan}, \ref{fig:gard_ex01_dmil}, \ref{fig:gard_ex01_ps} and~\ref{fig:gard_ex01_our}) depicts an area mainly characterized by tree crops, urban area and forest. Here, we highlight three representative zones on which classification differences are more evident. From the top to the bottom, the first two circles point out a field characterized by tree crops and forest zone respectively. On these two zones, we can observe that both $DMIL$ and $CNN_{PS}$ present confusion between these two classes and do not preserve the geometry of the scene. Conversely, we can observe that \method{} supplies a better (and more homogeneous) characterization of the two zones reducing confusion between the two classes and more correctly detecting parcel borders. The first zone highlighted in this example also involves an urban area. We can note that \method{} provides a more homogeneous classification of this zone w.r.t. the other two approaches that make some confusion between \textit{urban areas} and \textit{other crops} classes.

The second example (Figures~\ref{fig:gard_ex02_pan}, \ref{fig:gard_ex02_dmil}, \ref{fig:gard_ex02_ps} and~\ref{fig:gard_ex02_our}) represents a rural area mainly characterized by different crop types. Also in this case we highlight three zones to pinpoint the differences among the deep learning approaches. From the top to the bottom, the first focus is on a vineyard field. $DMIL$ and $CNN_{PS}$ have some issues to correctly assign the \textit{vineyard} class to the entire field making confusion among \textit{Tree crops} and \textit{Other Crops}. This is not the case for \method{} that provides a more correct delimitation compared to the correct one. The other two zones pointed out in this example involve an urban area and a forest field. We can observe that, also in this case, \method{} shows better performance on both \textit{Urban Areas} and \textit{Forest} classes than the other approaches.

The third example (Figures~\ref{fig:gard_ex03_pan}, \ref{fig:gard_ex03_dmil}, \ref{fig:gard_ex03_ps} and~\ref{fig:gard_ex03_our}) involves a wetland area. Here, we can clearly observe that the first two approaches ($DMIL$ and $CNN_{PS}$) have serious issues to recognize non water area and they tend to overestimate the water class. Conversely, \method{} achieves better performance to discriminate between \textit{water} and other classes. 

On this study area, \method{} seems to be more effective on some particular classes like \textit{Tree Crops}, \textit{Forest} and \textit{Urban areas}. These results are consistent with those reported in Table~\ref{tab:PerClass_fm_gard}. 
Considering a more fine visual inspection of the land cover maps, we can observe that the land cover map produced by $CNN_{PS}$ shows some horizontal strip artifact evident on the \textit{Tree Crops} class (orange color). $CNN_{PS}$ exhibits similar artifacts also on the second example. 

This behavior is not shared by the other approaches, which probably mean that such artifacts are due to some slight radiometric inconsistency of the pansharpened source. %Both $DMIL$ and \method{} take as input information at the original resolutions while $CNN_{PS}$ does not. 
%Considering $DMIL$ and \method{}, the former leverages internal upsampling (deconvolution operation inside the network~\cite{VolpiT17,AudebertSL16,NohHH15}) and it ignores the spatial information carried out by the MS image while the latter only applies convolutional operations (without introducing supplementary unnecessary parameters) and it also exploits the spatial information supplied by the MS as well as the one supplied by the PAN image. Leveraging the spatial information of both information sources allows \method{} to introduce a kind of spatial regularization that avoid strip behavior and produce more coherent and smoothed land cover maps. 

%Considering again the $CNN_{PS}$ approach, pansharpening artifacts influence the patch-based classification provided by $CNN_{PS}$. On the other hand, considering information at different resolution allows $DMIL$ and \method{} to introduce a kind of spatial regularization to avoid strip behavior and producing more coherent and smoothed land cover maps.% [KO] Again, due to resampling preprocess step, I think we could not say that DMIL use information at diffrent spatial resolution...

\begin{table*}[!ht]

\centering
\begin{tabular}{cccc}
\textit{VHSR Image} & \textbf{$DMIL$} & \textbf{$CNN_{PS}$} & \textbf{ \method } \\ 
\subfloat[\label{fig:reunion_ex01_pan}] {\includegraphics[width=.22\linewidth]{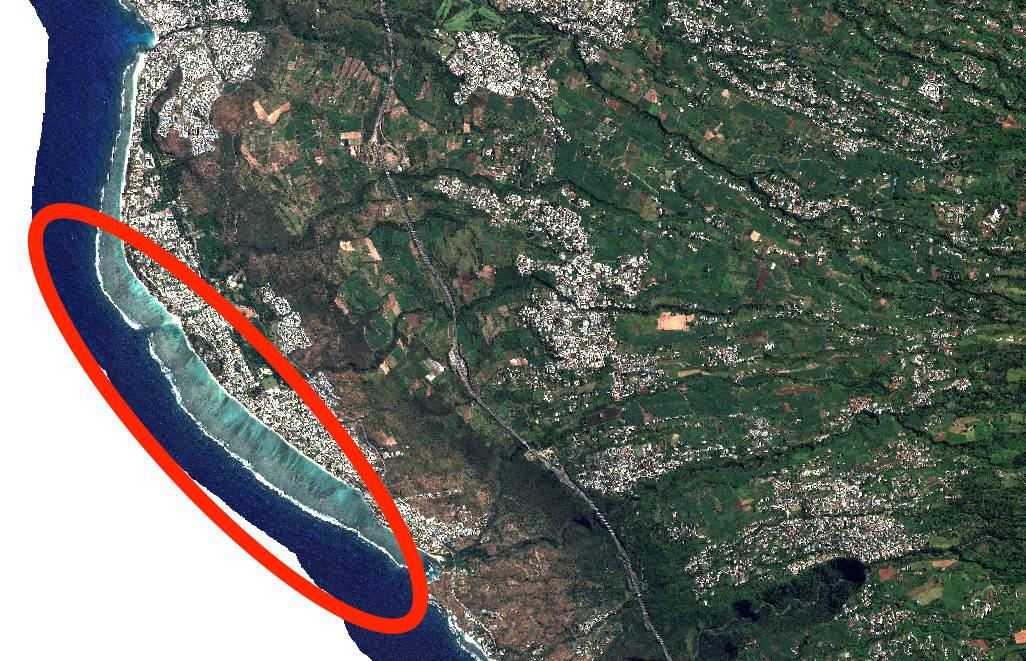} }
&
\subfloat[\label{fig:reunion_ex01_dmil}] {\includegraphics[width=.22\linewidth]{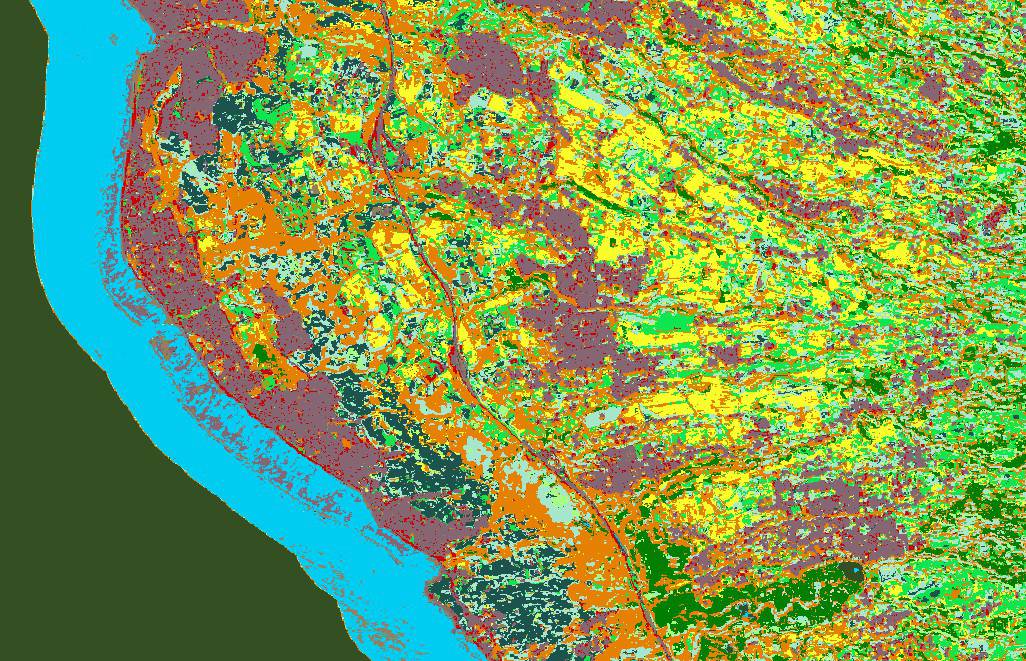} }
&
\subfloat[\label{fig:reunion_ex01_ps}] {\includegraphics[width=.22\linewidth]{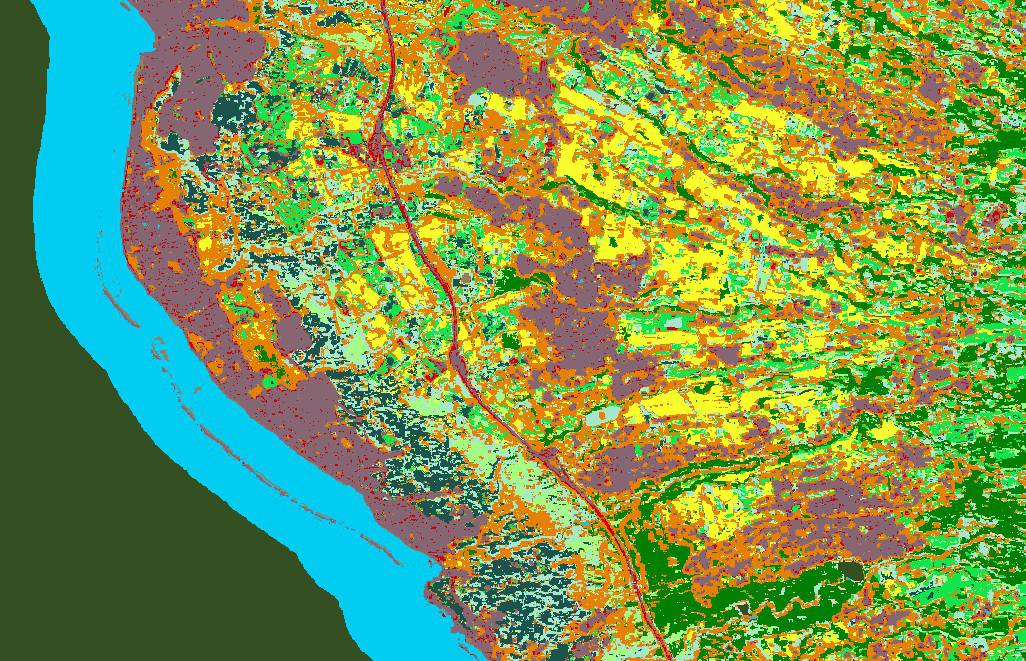} }
&
\subfloat[\label{fig:reunion_ex01_our}] {\includegraphics[width=.22\linewidth]{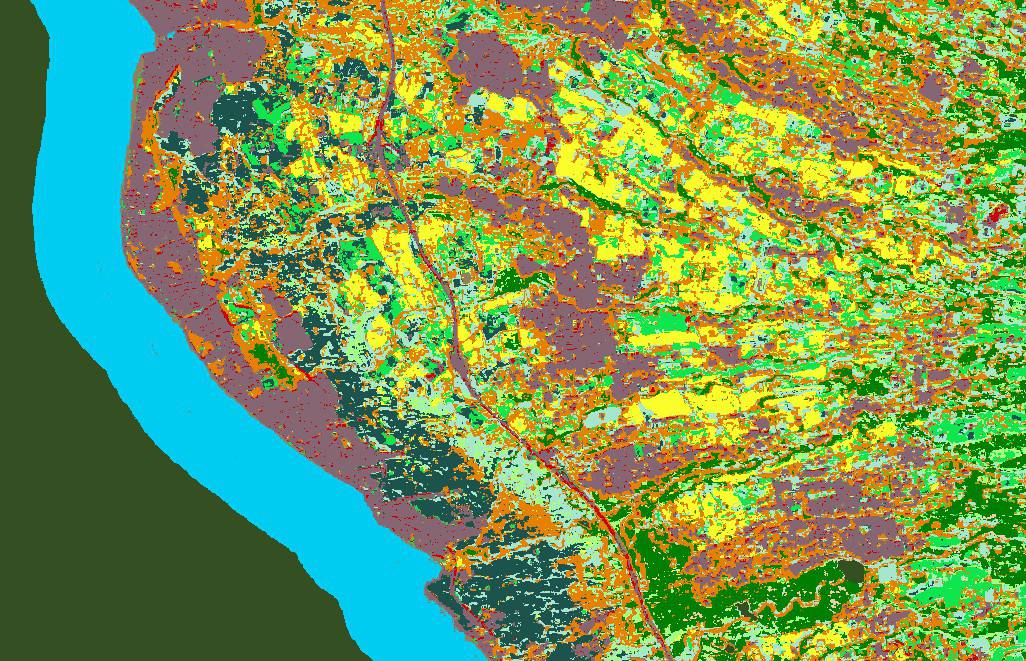} }
\\  
\subfloat[\label{fig:reunion_ex02_pan}] {\includegraphics[width=.22\linewidth]{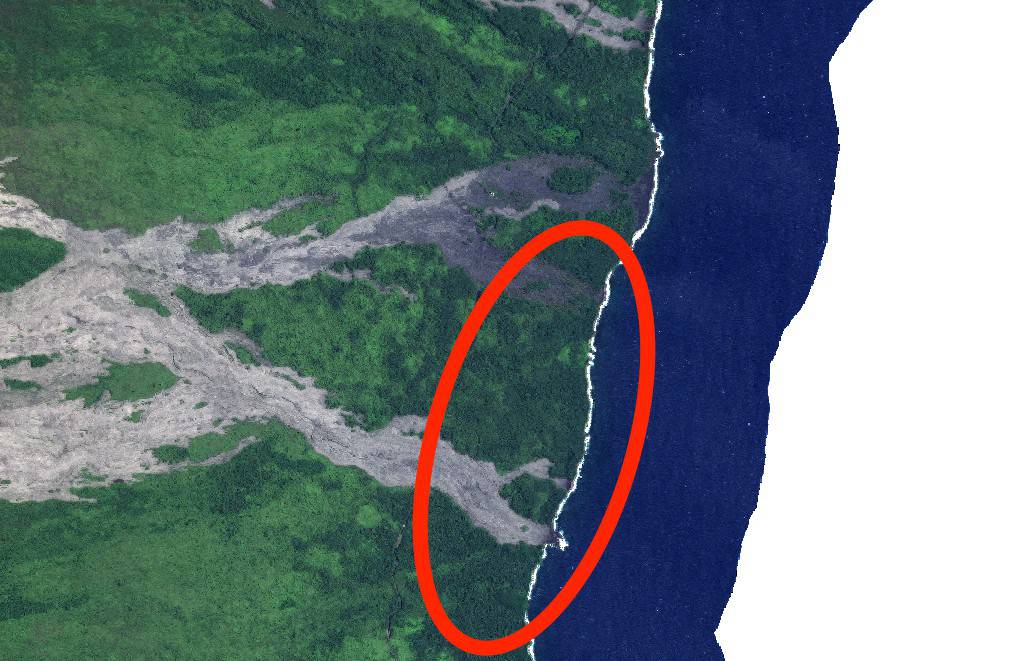} }
&
\subfloat[\label{fig:reunion_ex02_dmil}] {\includegraphics[width=.22\linewidth]{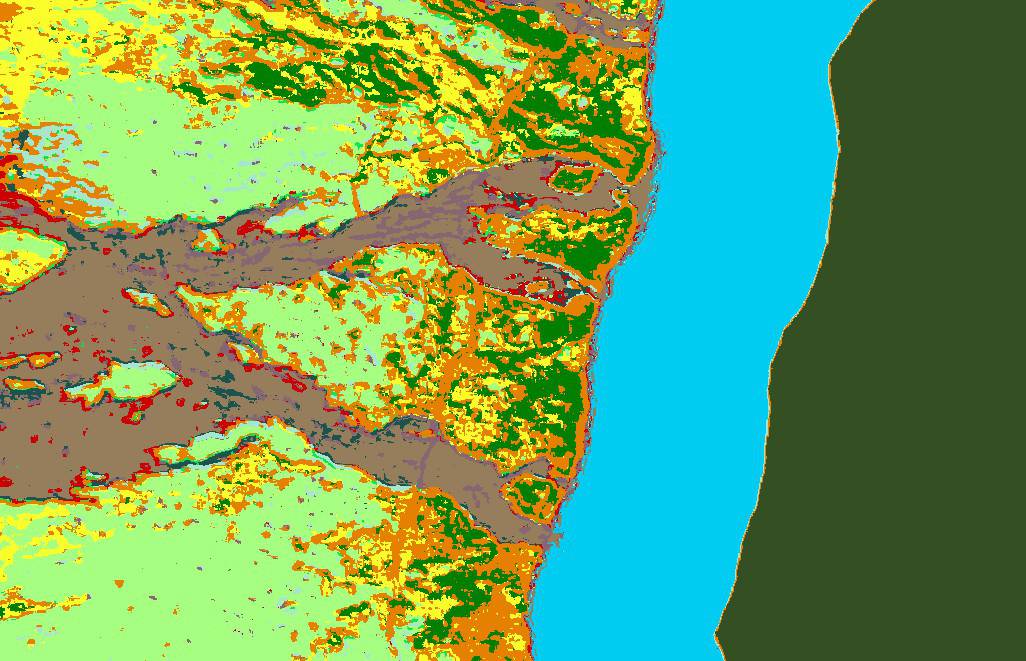} }
&
\subfloat[\label{fig:reunion_ex02_ps}] {\includegraphics[width=.22\linewidth]{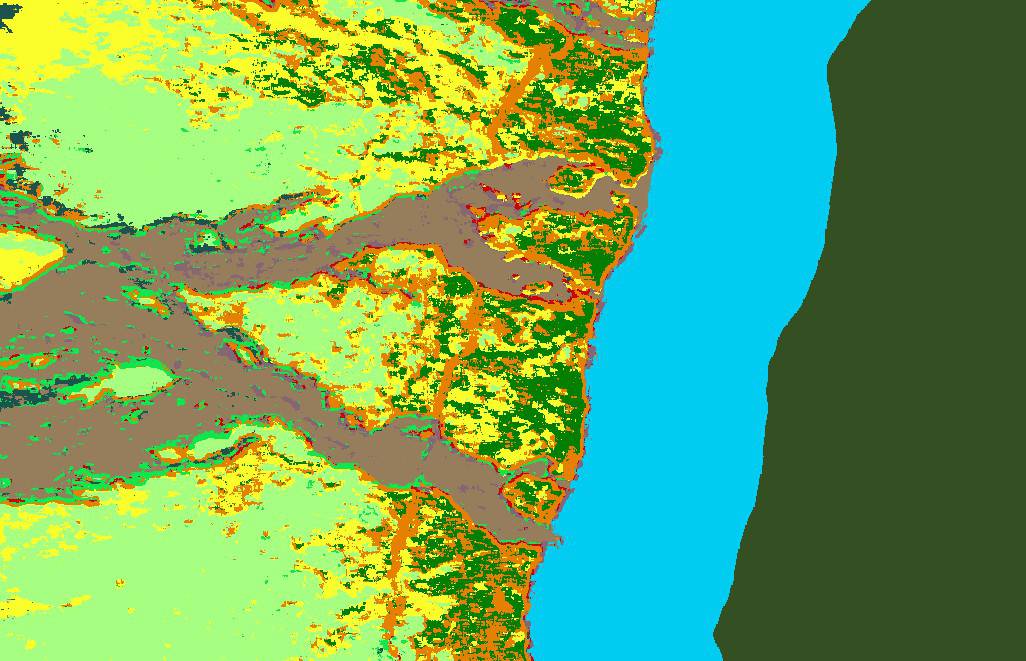} }
&
\subfloat[\label{fig:reunion_ex02_our}] {\includegraphics[width=.22\linewidth]{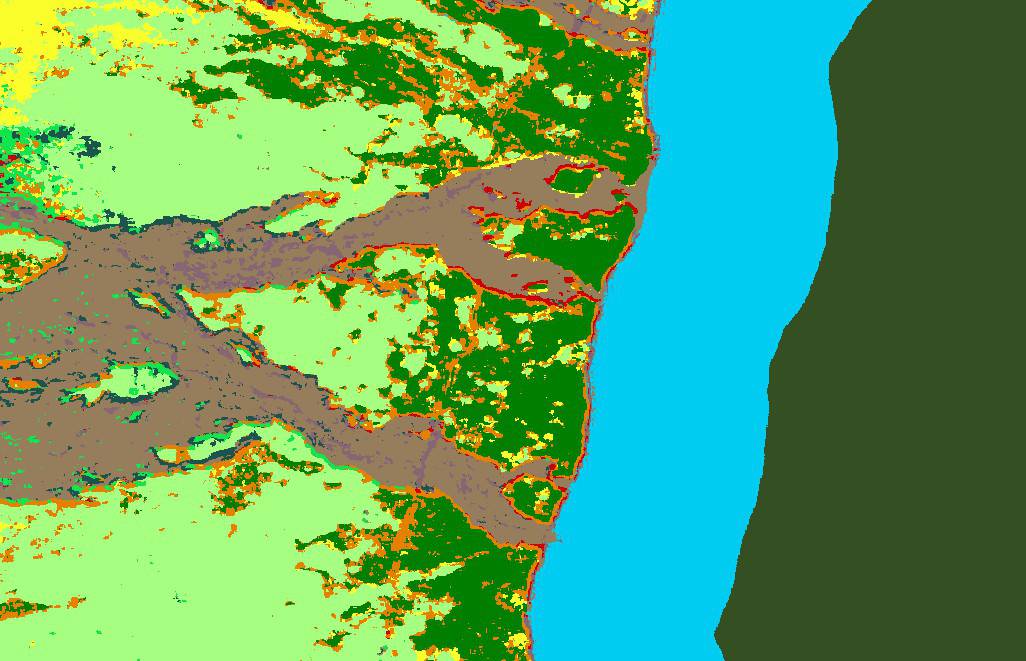} }
\\ 
\subfloat[\label{fig:reunion_ex03_pan}] {\includegraphics[width=.22\linewidth]{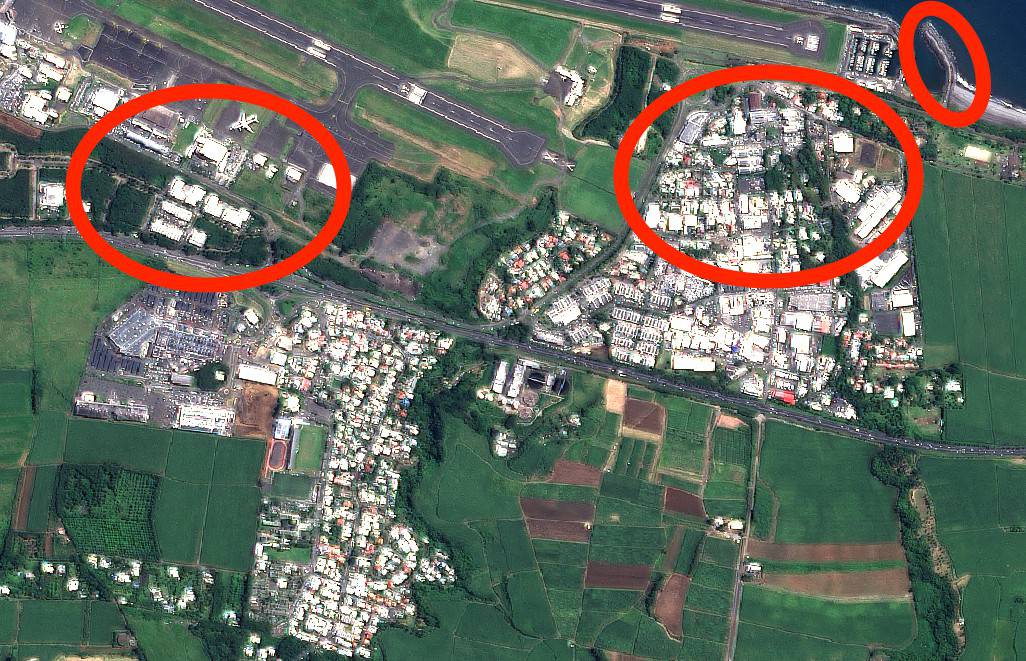} }
&
\subfloat[\label{fig:reunion_ex03_dmil}] {\includegraphics[width=.22\linewidth]{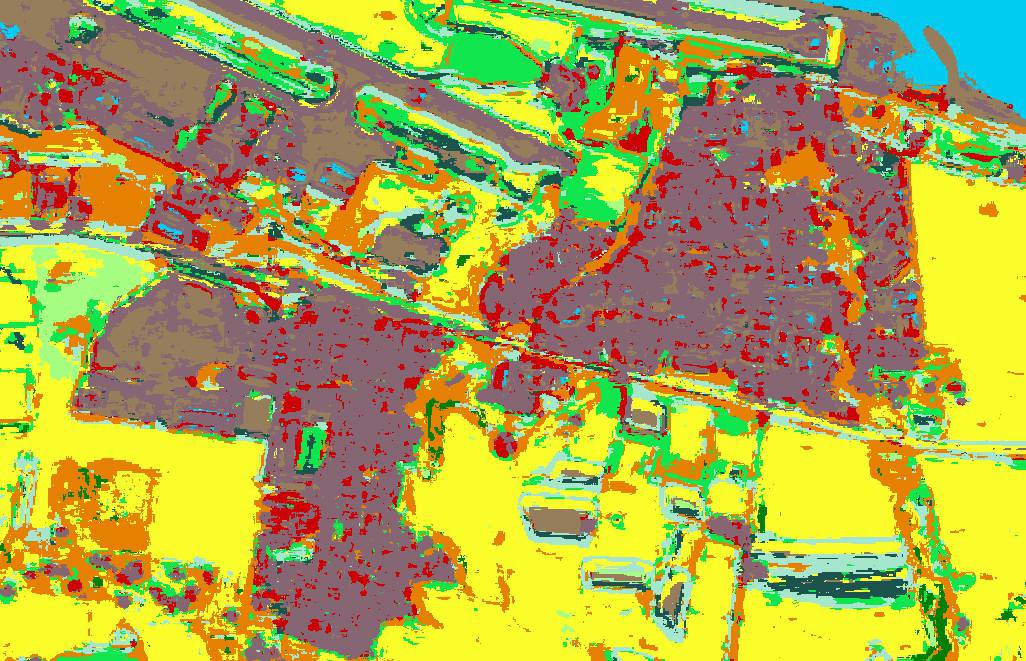} }
&
\subfloat[\label{fig:reunion_ex03_ps}] {\includegraphics[width=.22\linewidth]{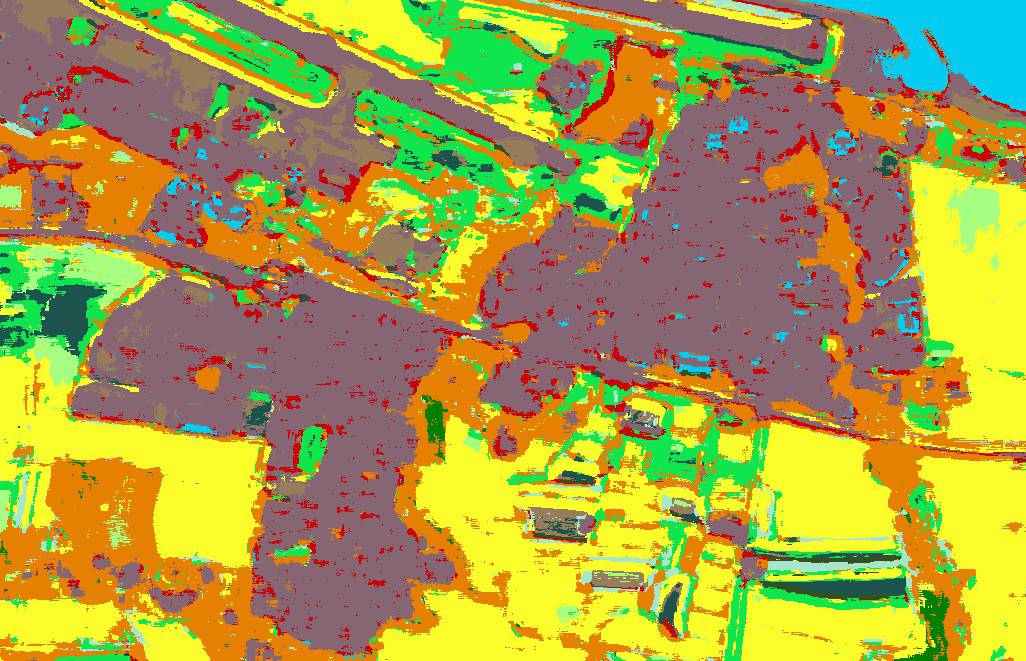} }
&
\subfloat[\label{fig:reunion_ex03_our}] {\includegraphics[width=.22\linewidth]{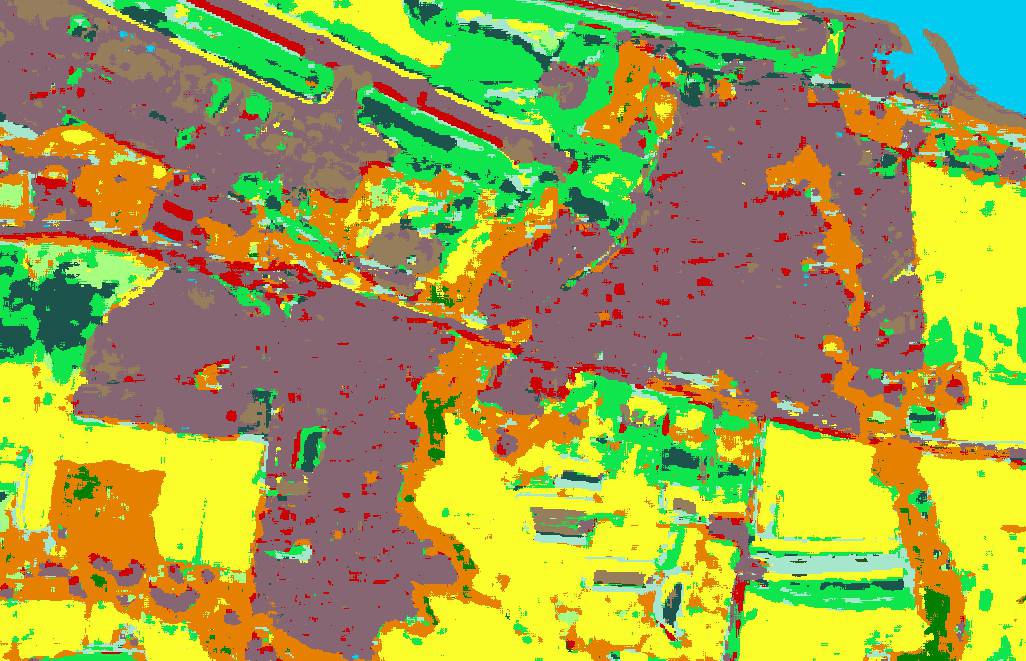} }
\\ 
\multicolumn{4}{c}{ \includegraphics[width=.8\textwidth]{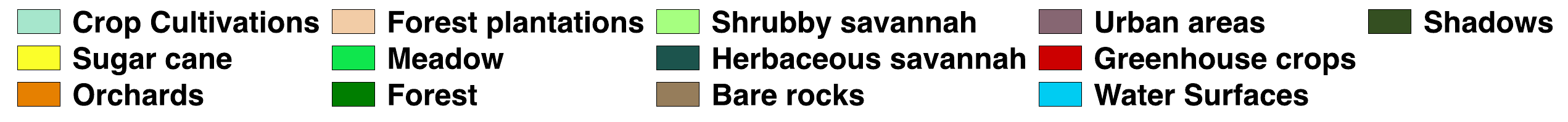}  }

\end{tabular}
\caption{table caption REUNION \label{tab:reunion_examples}}
\end{table*}

Concerning the \textit{Reunion Island} dataset, the first example (Figures~\ref{fig:reunion_ex01_pan}, \ref{fig:reunion_ex01_dmil}, \ref{fig:reunion_ex01_ps} and~\ref{fig:reunion_ex01_our}) depicts a coastal area of the West Coast of the Island. Here, we highlight a zone that is characterized by an underwater coral reef. $DMIL$ and $CNN_{PS}$ have some troubles to classify this zone like a water, more in detail they made confusion between the true class and the \textit{Bare Rocks} and \textit{Urban areas} classes. Conversely, \method{} does not have any issue with this point and it supplies a coherent water classification.

The second example (Figures~\ref{fig:reunion_ex02_pan}, \ref{fig:reunion_ex02_dmil}, \ref{fig:reunion_ex02_ps} and~\ref{fig:reunion_ex02_our}) represents a zone mainly characterized by forest. In this case, both $DMIL$ and $CNN_{PS}$ provide a noisy classification mixing \textit{Forest} with \textit{Sugar Cane} and \textit{Orchards}.
Conversely, when we analyze the land cover map produced by \method{} (Figure~\ref{fig:reunion_ex02_our}), we observe that the \textit{Forest} classification is more spatially homogeneous and consistent with the ground truth available in the corresponding VHSR image.

The third and last example related to the \textit{Reunion Island} dataset is supplied in Figures~\ref{fig:reunion_ex03_pan}, \ref{fig:reunion_ex03_dmil}, \ref{fig:reunion_ex03_ps} and~\ref{fig:reunion_ex03_our}. This area is mainly characterized by an urban settlement surrounded by some agricultural plots. The three zones highlighted in this example involve zones belonging to \textit{Urban Areas} and \textit{Bare rocks}. Comparing the maps provided by $DMIL$ and $CNN_{PS}$ w.r.t. the one provided by \method{}, we can note that the formers made more confusion between \textit{Urban Areas}/\textit{Bare rocks} and \textit{Water Surfaces} than the latter.
$DMIL$ and $CNN_{PS}$ tend to overestimate the prediction of the \textit{Water Surfaces} class. This phenomenon is more remarkable on the land cover map provided by $CNN_{PS}$ than the one supplied by $DMIL$. On the other hand, our approach has a more precise behavior on such classes and, as well as $DMIL$, it exploits the low-resolution information (Multi-Spectral bands) to regularize its spatial prediction.

Also on this study site \method{} exhibits a satisfactory behavior considering the competing approaches. Similarly to the analysis performed on the \textit{Gard} study site, results are consistent with those reported in Table~\ref{tab:PerClass_fm_reunion}. Conversely to what was proposed in $DMIL$, the joint utilization of spatial and spectral information of the MS and PAN images at their native resolution, without any intermediate upsampling step, provide useful regularization decreasing, at the same time, the confusion between land cover classes (i.e. \textit{Urban Areas} vs \textit{Water Surfaces} and \textit{Forest} vs \textit{Orchards}/\textit{Sugar Cane}). 

\section{Conclusion}
\label{sec:conclu}
In this paper, a novel Deep Learning architecture to fuse PAN and MS imagery for land cover mapping has been proposed. The approach, named \method{}, exploits Multi Spectral and Panchromatic information at their native resolutions. The architecture is composed of two branches, one for the PAN and one for the MS source. The final land cover mapping is achieved by concatenating the features extracted by each branch.% and use them to perform the final land cover mapping. 
The framework is learned end-to-end from scratch.

The evaluation on two real-world study sites has shown that \method{} achieves better quantitative and qualitative results than recent classification methods for optical VHSR images. In addition, the visual inspection of the land cover maps has underlined the effectiveness of our strategy and it advocates the use of both spatial and spectral information, at their native resolution, coming from PAN and MS imagery.

As future work, we plan to extend the approach on other optical Remote Sensing images, for instance dealing with classification on Sentinel-2 satellite images where the spectral information is available at different resolutions.

\section{Acknowledgements}
This work was supported by the French National Research Agency under the Investments for the Future Program, referred as ANR-16-CONV-0004 (DigitAg) and the GEOSUD project with reference ANR-10-EQPX-20, as well as from the financial contribution from the French Ministry of agriculture "Agricultural and Rural Development" trust account. This work also used an image acquired under the CNES Kalideos scheme (La R\'{e}union site).

%\subsection*{Annexe}
%Merci de votre participation.
%\bibliography{m3fusion}
\bibliography{PMS}
\bibliographystyle{IEEEtran}
\end{document}